\documentclass[journal,twoside,web]{ieeecolor}
\usepackage{etoolbox}
\makeatletter
\@ifundefined{color@begingroup}%
{\let\color@begingroup\relax
	\let\color@endgroup\relax}{}%
\def\fix@ieeecolor@hbox#1{%
	\hbox{\color@begingroup#1\color@endgroup}}
\patchcmd\@makecaption{\hbox}{\fix@ieeecolor@hbox}{}{\FAILED}
\patchcmd\@makecaption{\hbox}{\fix@ieeecolor@hbox}{}{\FAILED}
\usepackage{tmi}
\usepackage{cite}
\usepackage{amsmath,amssymb,amsfonts}
\usepackage{algorithmic}
\usepackage{graphicx}
\usepackage{textcomp}
\usepackage{bm}
\usepackage{diagbox} %插表格用的宏包
\usepackage{booktabs} %插表格用的宏包调整表格线与上下内容的间隔
\usepackage{multirow}
\usepackage{url}
\usepackage{pifont}
\usepackage{bbding}
\usepackage{cases}
\usepackage{subfigure}
\usepackage{graphicx}
\usepackage{subfig}
\usepackage[capitalize]{cleveref}
\def\BibTeX{{\rm B\kern-.05em{\sc i\kern-.025em b}\kern-.08em
    T\kern-.1667em\lower.7ex\hbox{E}\kern-.125emX}}
\begin{document}
\title{Prompting Lipschitz-constrained network for multiple-in-one sparse-view CT reconstruction}
\author{Baoshun Shi, \IEEEmembership{Senior Member, IEEE}, Ke Jiang, Qiusheng Lian, Xinran Yu, and Huazhu Fu, \IEEEmembership{Senior Member, IEEE}
\thanks{This work was supported by the National Natural Science Foundation of China under Grants 62371414 and 62571473, by the Hebei Natural Science Foundation under Grant F2025203070, and by the Hebei Key Laboratory Project under Grants 202250701010046. The authors thank the anonymous reviewers for constructive suggestions. \textit{(Baoshun Shi and Ke Jiang contributed equally to this work.)(Corresponding authors: Baoshun Shi and Qiusheng Lian.)}}
\thanks{Baoshun Shi, Ke Jiang, and Qiusheng Lian are with the School of Information Science and Engineering, Yanshan University, Qinhuang Dao, 066004, Hebei province, China (e-mail: shibaoshun@ysu.edu.cn, lianqs@ysu.edu.cn).}
\thanks{Xinran Yu is with the Beijing Advanced Innovation Center for Imaging Technology, Capital Normal University, Beijing 100048, China.}
\thanks{Huazhu Fu is with the Institute of High Performance Computing (IHPC), Agency for Science, Technology and Research (A*STAR), Singapore 138632.}}
\maketitle

\begin{abstract}	
Despite significant advancements in deep learning-based sparse-view computed tomography (SVCT) reconstruction algorithms, these methods still encounter two primary limitations: (\textit{\romannumeral1}) It is challenging to explicitly prove that the prior networks of deep unfolding algorithms satisfy Lipschitz constraints due to their empirically designed nature. (\textit{\romannumeral2}) The substantial storage costs of training a separate model for each setting in the case of multiple views hinder practical clinical applications. To address these issues, we elaborate an explicitly provable Lipschitz-constrained network, dubbed LipNet, and integrate an explicit prompt module to provide discriminative knowledge of different sparse sampling settings, enabling the treatment of multiple sparse view configurations within a single model. Furthermore, we develop a storage-saving deep unfolding framework for multiple-in-one SVCT reconstruction, termed PromptCT, which embeds LipNet as its prior network to ensure the convergence of its corresponding iterative algorithm. In simulated and real data experiments, PromptCT outperforms benchmark reconstruction algorithms in multiple-in-one SVCT reconstruction, achieving higher-quality reconstructions with lower storage costs. On the theoretical side, we explicitly demonstrate that LipNet satisfies boundary property, further proving its Lipschitz continuity and subsequently analyzing the convergence of the proposed iterative algorithms. The data and code are publicly available at \url{https://github.com/shibaoshun/PromptCT}.
\end{abstract}


\section{Introduction}
\label{sec:introduction}
\IEEEPARstart{X}{-ray} computed tomography (CT) is a crucial diagnostic technology in medical imaging. However, prolonged or excessive exposure to X-ray radiation may pose potential risks of radiation-related diseases\cite{1}. Sparse-view CT (SVCT) serves as an effective solution, reducing human radiation exposure by acquiring partial projection data through equidistant sampling over the full scanning range\cite{3,wu1}. This approach not only shortens scanning time but also mitigates motion artifacts caused by shaking, heartbeat, and respiration\cite{4}. Nevertheless, the absence of projection data at certain angles may result in severe global streak artifacts in CT images reconstructed using filtered back-projection (FBP), potentially compromising critical tissue details and hindering clinical diagnosis. Effectively reconstructing high-quality CT images from sparse-view projection data remains a significant challenge \cite{wu2}.\par
\begin{figure*}[!ht]
	\centering
	\subfigure[]{
		\begin{minipage}[t]{0.5\textwidth}
			\centering
			\includegraphics[width=9cm]{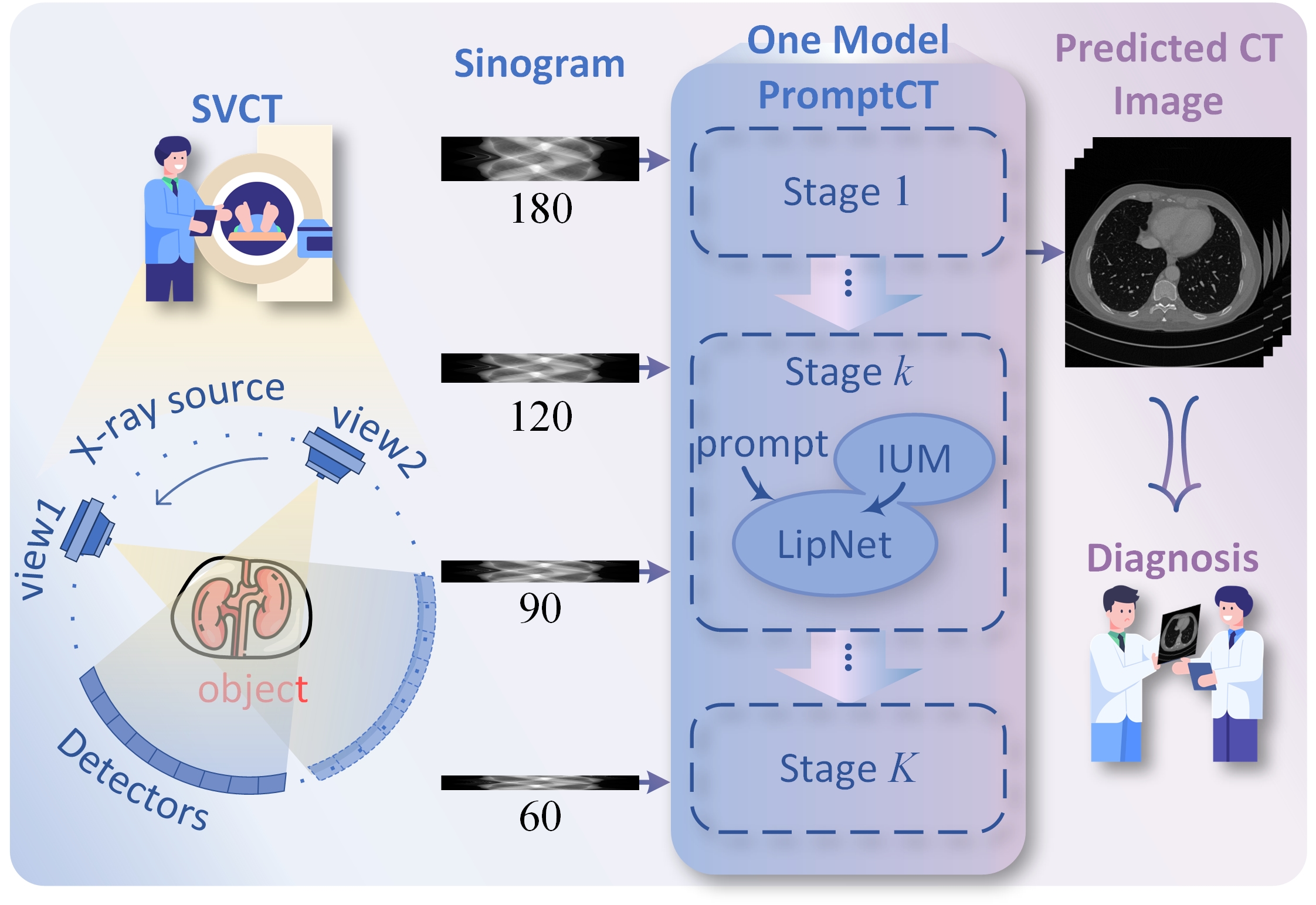}
		\end{minipage}%
	}\hspace{-7mm}
	\subfigure[]{
		\begin{minipage}[t]{0.5\textwidth}
			\centering
			\includegraphics[width=7.75cm]{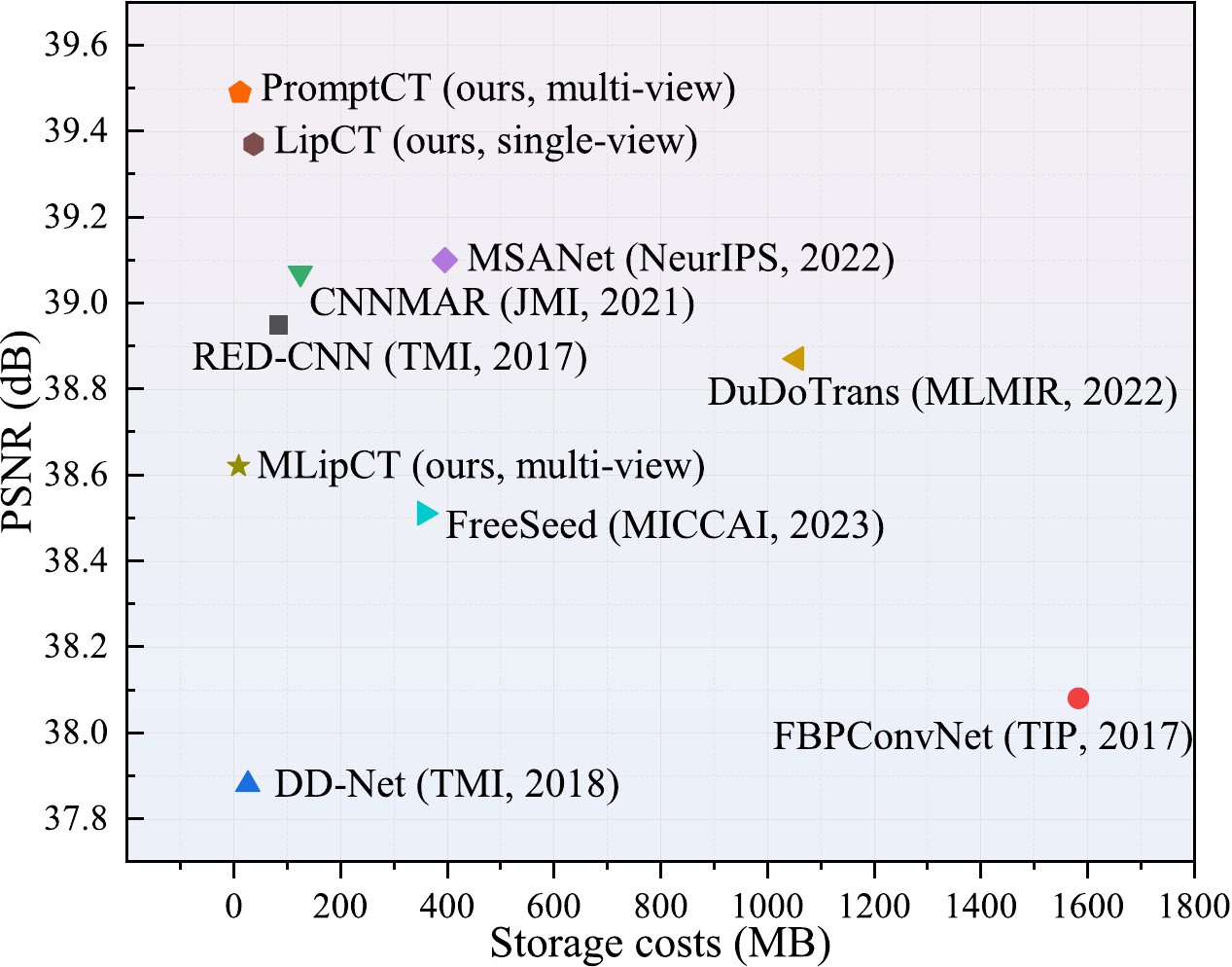}
		\end{minipage}%
	}\vspace{-0.3cm}
    \caption{(a) The proposed deep unfolding PromptCT enables SVCT reconstruction across varying numbers of projection data, e.g., 60, 90, 120, and 180, using a single model to aid in medical diagnosis and treatment, where each stage incorporates the image update module (IUM) and the prompt-based LipNet that satisfies the Lipschitz constraint. (b) We compare the average PSNR (dB) and storage cost (MB) of the proposed SVCT method and the existing SVCT methods under four sampling views. It reveals that the proposed strategy significantly saves storage costs while improving reconstruction performance. Notably, MLipCT is the multi-view model of the backbone architecture, i.e., LipCT, without explicit prompt module, and PromptCT is the multi-view model with explicit prompt module.}
    \vspace{-0.4cm}
    \label{fig1}
\end{figure*}
Traditional iterative algorithms typically tackle the SVCT reconstruction problem by imposing various hand-crafted priors. Due to the insufficient characterization of image features by hand-crafted priors and numerous iterations required for algorithm convergence, these algorithms suffer from low-quality reconstruction and heavy computational burden. Inspired by the significant success of deep learning (DL) in medical image reconstruction, deep neural networks (DNNs) have been applied to the SVCT reconstruction task\cite{data2}. Despite the high-quality reconstructions achieved by these data-driven methods, the model architectures of DNNs typically lack interpretability, thereby hindering further theoretical analysis\cite{29,30,Liu,SPL}.\par
Recently, a promising research direction for interpretable networks is the "iterative theory + deep learning" scheme, which can be divided into plug-and-play (PnP) reconstruction methods\cite{17,19} and deep unfolding reconstruction methods\cite{35}. On the theoretical side, if the PnP iterative algorithms are unrolled, the convergence analysis of the PnP algorithms can be transferred to analyze the stability of the corresponding unrolled algorithms\cite{Yang}, i.e., the convergent PnP algorithms lead to stable unrolled algorithms, with their performance becoming increasingly stable as the stage increases\cite{Yang}. However, the theoretical analysis of existing PnP or deep unfolding methods still faces several limitations\cite{14,15}. In particular, the selection and implementation of denoisers play a critical role in PnP reconstruction algorithm convergence\cite{20,21,43}. In fact, to ensure algorithm convergence, the use of bounded denoisers necessitates that the gradients of data fidelity terms are bounded\cite{16}. However, incomplete data, noise interference, and computational complexity often render closed-form solutions to optimization problems infeasible in most medical imaging tasks, potentially limiting the direct application of bounded denoisers in proving algorithm convergence. On the other hand, traditional deep unfolding algorithms often rely on empirically designed prior networks, which perform well in practical applications but are difficult to explicitly prove satisfying boundary properties or Lipschitz conditions. These conditions are common prerequisites for ensuring algorithm convergence.\par
On the practical side, there exists a range of sparse-view sampling strategies tailored to specific clinical requirements\cite{6}. The distribution of artifacts in the reconstructed CT images varies under different sampling views or under-sampling rates. In recent years, existing DL-based SVCT reconstruction methods typically handle these sparse sampling configurations individually by training the separate model for each specific sparse-view setting\cite{8,9,10}. Although this "one-model-for-one-setting" approach demonstrates excellent experimental performance, it requires substantial storage costs, and the flexibility of the single model limits clinical applications. Inspired by prompt learning, all-in-one methods typically train a general model for various tasks, but they require extensive datasets\cite{39}. However, acquiring large amounts of paired CT data is impractical in medical imaging.\par
To tackle the aforementioned issues, we propose a prompting Lipschitz-constrained network for multiple-in-one SVCT reconstruction. This "one-model-for-multi-view" strategy aims to develop a universal multi-view model for multiple sampling views in the SVCT reconstruction task, thereby generating high-quality CT images to assist doctors in diagnosis. In theory, the proposed Lipschitz-constrained network, serving as a prior network, cleverly integrates the two theories of boundary property and Lipschitz constraint, which successfully circumvents the stringent restrictions on the data fidelity terms while ensuring the convergence of iterative algorithms. Particularly, we introduce sparse sampling masks as explicit prompts to develop a flexible single model that is trained to handle the SVCT reconstruction with different sampling ratios. The contributions of this paper are as follows:\par
$\bullet$ We propose an explicitly provable Lipschitz-constrained sparse representation model-driven network, termed LipNet, constructed from a deep unfolding sparse representation framework that satisfies boundary property. Within LipNet, we elaborate a constant-generating sub-network (CGNet) with an explicit prompt module to determine view-aware and spatial-variant thresholds.\par
$\bullet$ We exploit sinogram sampling masks as explicit prompts to distinguish view information from different sparse-view settings. Even using thousands of samples, the explicit prompt module can classify mixed feature information, enabling the reconstruction network to handle multiple sparse sampling settings within a single universal model.\par
$\bullet$ We devise the so-called CGNet composed of the shallow feature extraction module, swin Transformer block (STB), and spatial frequency block (SFB) to adaptively generate the proportional constants of thresholds by extracting local, regional, and global information from representation coefficients, thereby improving the reconstruction performance and representation ability.\par
$\bullet$ As depicted in Fig. \ref{fig1}, we construct a storage-saving deep unfolding network for multiple-in-one SVCT reconstruction, called PromptCT, which incorporates the proximal gradient descent technique for algorithm optimization. We replace the proximal operator in PromptCT with the well-designed LipNet, providing a clear working mechanism and convergence analysis. Comprehensive experiments, including synthetic-to-real generalization, finely substantiate that PromptCT achieves outstanding performance across various sparse-view settings using a multi-view model, surpassing existing SVCT reconstruction methods. On the theoretical side, we explicitly demonstrate that LipNet satisfies boundary property and Lipschitz continuous, and further prove the convergence of the proposed iterative algorithm.\par
The remainder of the manuscript is structured as follows. In Section II, we provide a brief overview of related works. Section III introduces the prompting Lipschitz-constrained network and its theoretical analysis. Section IV focuses on introducing the deep unfolding SVCT reconstruction network and its theoretical analysis. Section V showcases the experimental evaluations to validate the superiority of the proposed network. Section VI summarizes our work and outlines potential directions for future research.\par
\section{Related Work}
\label{sec:Related Work}
\subsection{SVCT reconstruction}
\label{subsec:Sparse-view CT reconstruction}
In general, existing DL-based SVCT reconstruction approaches can be roughly grouped into two categories, i.e., data-driven methods and model-driven methods\cite{23}. Specifically, data-driven SVCT reconstruction methods rely on a large number of input-output data pairs for training, directly learning the mapping relationship and feature representations of the reconstruction task from the data\cite{26,28}. Although these methods, which do not rely on prior knowledge or require modeling of physical processes, often exhibit strong flexibility, the black-box nature of DNNs typically results in insufficient model interpretability for these data-driven networks.\par
In recent years, many efforts have focused on incorporating the DNNs into the traditional iterative framework through PnP strategy\cite{34} or deep unfolding strategy\cite{3,35,36}, to improve the interpretability and representation ability of the network. In theory, common PnP frameworks combining iterative reconstruction algorithms with pre-trained Gaussian denoisers can achieve promising performance in a wide range of imaging tasks. However, the solutions provided by these frameworks remain suboptimal due to the non-adaptive deep Gaussian denoiser. In contrast, the deep unfolding method replaces these pre-trained deep denoisers of iterative algorithms with DNNs through end-to-end learning\cite{37,11}. These model-driven networks combine the interpretability of iterative algorithms with the powerful reconstruction ability of learning-based approaches.\par
\subsection{Prompt learning}
\label{subsec:Prompt learning}
Prompt learning\cite{38} is an emerging technique in the fields of natural language processing and machine learning, experiencing rapid development and garnering widespread attention. Initially, prompt learning involves studying how to introduce additional texts (i.e., prompts) as inputs to pre-trained large language models to obtain the desired outputs. With further research, prompt-based approaches aim to provide contextual information to the model for fine-tuning towards target tasks, enabling parameters to adapt the model more effectively, thus offering significant flexibility. In the field of image restoration, some works utilize learnable prompts to distinguish different image restoration tasks and demonstrate their effectiveness in low-level vision tasks\cite{39}. For example, Gao et al. \cite{10526271} employed degradation-specific information to dynamically guide restoration networks using learned prompts, while Zhang et al. \cite{10204072} proposed a two-stage approach that encompasses task-specific knowledge collection and component-oriented knowledge integration. Further advancements came from Park et al. \cite{10204770}, who employed adaptive discriminative filters, and Zhu et al.\cite{10204275}, who concentrated on learning both general and weather-specific features to improve performance in weather-degraded conditions. Learnable prompts require a large number of samples for effective feature discrimination, thus this strategy is not applicable in medical image reconstruction tasks since it is difficult to acquire massive CT images. Therefore, we propose using explicit prompts instead of learnable prompts to efficiently embed discriminative information into the data flow of the reconstruction network only using thousands of samples.\par
\begin{figure*}[!t]
	\centering
	\includegraphics[width=1\linewidth]{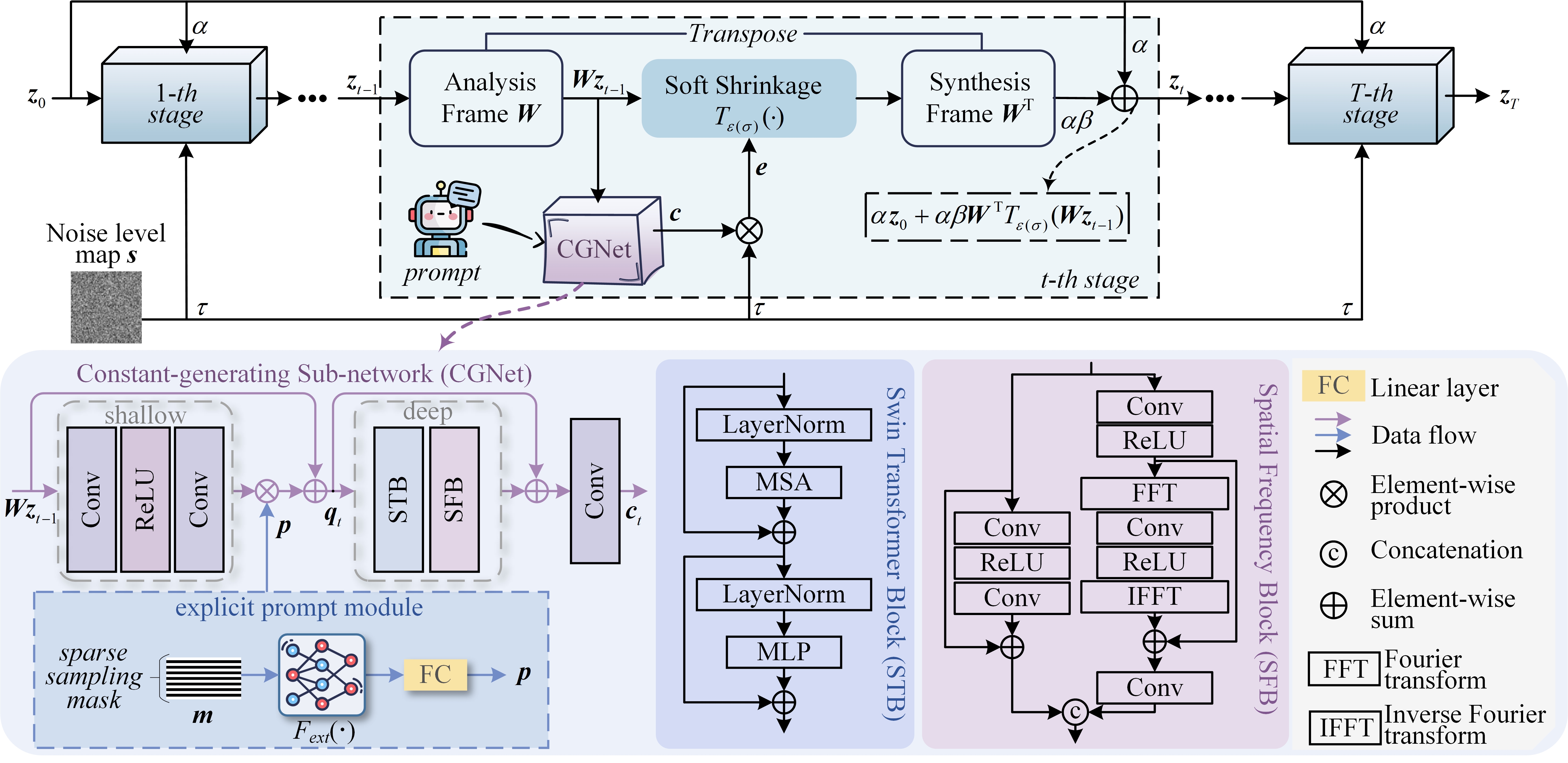}
	\caption{The network architecture of LipNet. Each stage of this network consists of an analysis frame $\bm{W}$, a soft shrinkage operation, a synthesis frame $\bm{W}^{\rm{T}}$, and a constant-generating sub-network (CGNet), forming a single-layer sparse representation model-driven architecture. The element-wise products of the generated proportional constants $\bm{c}$ and the elements in the input noisy map $\bm{s}$ are used as the thresholds $\bm{e}$ for shrinking frame coefficients. The network architecture of CGNet comprises the shallow feature extraction module, the deep feature extraction module, and the explicit prompt module, utilized for generating spatial-variant proportional constants. Among them, the shallow feature extraction contains two $3\times3$ convolutional layers, and the deep feature extraction consists of STB and SFB for extracting local, regional, and global feature information, respectively. In addition, we embed view prompts between these two feature extraction modules to guide the differentiation of multiple sparse sampling views.}
	\vspace{-0.3cm}
	\label{fig4}
\end{figure*}
\section{The proposed prompting Lipschitz-constrained network}
\label{sec:LipNet}
\subsection{LipNet: when bounded network meets Lipschitz-constrained network}
\label{subsec:meets}
The convergence proofs of the PnP algorithms can be transferred to analyze the stability of deep unfolding methods\cite{Yang}. Therefore, the properties of denoisers in PnP algorithms can be extended to the prior networks in deep unfolding algorithms. Based on this fact, we extend the bounded denoiser to the bounded network and generalize the concept of Lipschitz constraint to the Lipschitz-constrained network, both of which can serve as prior networks extended to deep unfolding algorithms, ensuring the convergence of their iterative algorithms. The definitions of bounded network and Lipschitz-constrained network are provided as follows.\par
\noindent\textbf{Definition 1 (Bounded network)}: The procedure of a DNN denoted as $\mathcal{D}_{B}(\cdot;\sigma)$ with a input parameter $\sigma$: $\mathbb{R}^{n} \to \mathbb{R}^{n}$ such that for any input $\bm{x}\in\mathbb{R}^{n}$, the following inequality holds
\begin{equation} 
	||\mathcal{D}_{B}(\bm{x};\sigma)-\bm{x}||_2^2\le\sigma^{2}C
	\label{e1}
\end{equation}
where $\sigma$ is the noise level and the universal constant $C$ is independent of $\sigma$. Under the condition of bounded networks, the diminishing noise level condition is crucial for the convergence analysis of the algorithm.\par
\noindent\textbf{Definition 2 (Lipschitz-constrained network)}: The procedure of a DNN denoted as $\mathcal{D}_{L}(\cdot)$ is Lipschitz continuous with a constant $L>0$ if
\begin{equation} 
	||\mathcal{D}_{L}(\bm{x}_1)-\mathcal{D}_{L}(\bm{x}_2)||_2\le L||\bm{x}_1-\bm{x}_2||_2
	\label{e2}
\end{equation}
for any input $\bm{x}_1,\bm{x}_2 \to \mathbb{R}^{n}$.\par
In PnP reconstruction algorithms, employing a bounded denoiser to demonstrate algorithm convergence requires ensuring that the gradient of the data fidelity term is bounded\cite{17,18}. However, the optimization problem containing the data fidelity term demands solving closed-form solutions, which poses practical challenges for complex imaging scenarios such as the SVCT reconstruction task. On the other hand, although imposing the assumption that denoisers satisfy Lipschitz constraints can guarantee the convergence of PnP reconstruction algorithms, explicitly proving that denoisers adhere to Lipschitz constraints remains a challenge. Some efforts focus on imposing constraints on filter weights, but these approaches reduce the representation ability of the network, thereby leading to a degradation in reconstruction performance\cite{20}. In this paper, we integrate the theories of boundary properties and Lipschitz constraints to construct a network that explicitly satisfies Lipschitz constraints by leveraging bounded networks. This approach effectively mitigates the constraints associated with the two major theories, thereby avoiding the need for closed-form solutions and making it more suitable for SVCT reconstruction tasks.\par 
As a popular representation model, the sparse representation model, which can promote the sparsity of images over some sparsifying frames, is commonly used to construct provable networks and has been proven effective in imaging fields \cite{17,9763013}. Specially, a frame $\bm{W}$ equipped with the tight property $\bm{W}^{\rm{T}}\bm{W}=\bm{I}$ is called a tight frame, whose inverse transform is its transpose. Tight frames can be roughly classified into analytical tight frames \cite{5413975} and data-driven tight frames \cite{CAI201489}. Imaging algorithms using analytical tight frames are often faster, but their reconstructions are inferior compared with those using data-driven tight frames. Data-driven tight frames, being adaptive to the data, can capture more structural information compared to analytical ones. In general, traditional tight frame learning methods often update the tight frames and their corresponding frame coefficients via a time-consuming alternating optimization strategy, which hinders the incorporation of modern deep learning techniques \cite{19M1298524}. Existing bounded denoising networks are predominantly constructed using tight frames\cite{18,19}, which must adhere to the strict constraint $\bm{W}^{\rm{T}}\bm{W}=\bm{I}$. This requirement imposes substantial limitations on their flexibility and representation ability.\par
To address this limitation, we cleverly relax the strict tight frame constraint and propose a novel approach that constructs bounded networks using general sparsifying frames instead of tight frames. Specifically, we propose to learn both the analysis frame $\bm{W}$ and the synthesis frame $\bm{W}^{\rm{T}}$ in an end-to-end supervised learning manner. To further enhance the representation ability of these sparsifying frames, we design a CGNet that not only improves their adaptability but also eliminates the need for laborious threshold tuning. This approach leverages the benefits of frame-based representation while overcoming the drawbacks of tight frame constraints, providing a more flexible and effective framework for image reconstruction tasks. Furthermore, instead of strict sparsity, we explore approximate sparsity, which is widely applied in the sparsifying transform learning\cite{wen1,wen2}. Assuming that the input image is approximately sparse under the learnable sparsifying frame $\bm{W}$, the optimization problem for filtering can be formulated as
\begin{equation} 
	\mathop{\min}\limits_{\bm{z}}\frac{1}{2}||\bm{z}_{0}-\bm{z}||_2^2+\mu||\bm{W}\bm{z}||_{1}
	\label{eq:3}
\end{equation}
where $\bm{z}_{0}$ is the underlying image, $\bm{z}$ is the introduced auxiliary variable, $\mu$ denotes the trade-off parameter, $||\cdot||_1$ represents the ${l}_1$ norm, and $\bm{W}$ is the learnable sparsifying frame or filter. Notably, compared with previous tight frame networks\cite{17,18,43}, our filter does not require the tight constraint. By leveraging an auxiliary variable $\bm{u}$, Eqn. (\ref{eq:3}) can be rewritten as
\begin{equation} 
	\mathop{\min}\limits_{\bm{z},\bm{u}}\frac{1}{2}||\bm{z}_{0}-\bm{z}||_2^2+\mu||\bm{u}||_1,\ s.t.\ \bm{W}\bm{z}=\bm{u}.
	\label{eq:4}
\end{equation}
Subsequently, we solve the problem defined in Eqn. (\ref{eq:4}) by using the Half Quadratic Splitting (HQS) solver. Thus, a sequence of individual subproblems emerges (for the $t$-th iteration):
\begin{equation} 
	\bm{z}_{t}=\mathop{\arg\min}\limits_{\bm{z}}\left\{\frac{1}{2}||\bm{z}_{0}-\bm{z}||_2^2+\frac{\beta}{2}||\bm{Wz}-\bm{u}_{t-1}||_2^2\right\}
	\label{eq:5}
\end{equation}
and
\begin{equation} 
	\bm{u}_{t}=\mathop{\arg\min}\limits_{\bm{u}}\left\{\frac{\beta}{2}||\bm{Wz}_{t}-\bm{u}||_2^2+\mu||\bm{u}||_1\right\}
	\label{eq:6}
\end{equation}
where $\beta$ stands for a penalty parameter. The iterative solutions of subproblems for $\bm{z}$ and $\bm{u}$ are defined as
\begin{equation} 
	\bm{z}_{t}=\frac{\bm{z}_{0}+\beta\bm{W}^{\rm{T}}\bm{u}_{t-1}}{\bm{I}+\beta\bm{W}^{\rm{T}}\bm{W}}
	\label{eq:7}
\end{equation}
and
\begin{equation} 
	\bm{u}_{t}=T_{\varepsilon(\sigma)}(\bm{W}\bm{z}_{t})
	\label{eq:8}
\end{equation}
where $T_{\varepsilon(\sigma)}(\cdot)$ is the soft thresholding operator $soft_{\varepsilon(\sigma)}=sign(u)\max(|u|-\varepsilon(\sigma),0)$ and the threshold $\varepsilon(\sigma)$ is correlated with the noise strandard deviation $\sigma$. Based on Eqn. (\ref{eq:8}), we have $\bm{u}_{t-1}=T_{\varepsilon(\sigma)}(\bm{W}\bm{z}_{t-1})$. Substituting this equation into Eqn. (\ref{eq:7}), we can recast Eqn. (\ref{eq:7}) as
\begin{equation} 
	\bm{z}_{t}=\frac{\bm{z}_{0}+\beta\bm{W}^{\rm{T}}T_{\varepsilon(\sigma)}(\bm{W}\bm{z}_{t-1})}{\bm{I}+\beta\bm{W}^{\rm{T}}\bm{W}}.
	\label{eq:9}
\end{equation}
Define $\alpha=\frac{1}{\bm{I}+\beta\bm{W}^{\rm{T}}\bm{W}}$ as a constant less than 1, the proposed iterative updated rule is summarized as follows
\begin{equation} 
	\bm{z}_{t}=\alpha\bm{z}_{0}+\alpha\beta\bm{W}^{\rm{T}}T_{\varepsilon(\sigma)}(\bm{W}\bm{z}_{t-1}).
	\label{eq:10}
\end{equation}
In summary, the problem defined in Eqn. (\ref{eq:3}) can be solved via iterating the inversion and filter steps alternatively. Considering the aforementioned solving process, the proposed LipNet incorporates a parameter $\sigma$ and is modeled as a function $\mathcal{D}_{\theta}(\cdot;\sigma)$, defined as $\bm{z}_{T}=\mathcal{D}_{\theta}(\bm{z}_{0};\sigma)$, where $\theta$ represents the set of learnable parameters within this DNN. Based on the interpretability of the sparse representation model and the strong learning ability of DNNs, the proposed LipNet is constructed as a deep unfolding network in the model- and data-driven manner by unfolding the iteration defined in Eqn. (\ref{eq:10}). The architecture of our proposed LipNet is illustrated in Fig. \ref{fig4}. In LipNet, each stage corresponds rigorously to each iteration of the HQS algorithm. Moreover, in order to enhance the representation ability of the network, we elaborate a constant-generating sub-network based on explicit prompts to generate adaptive thresholds, and the concrete network architecture is introduced in Section \ref{subsec:constant network}.
\subsection{The prompt-based constant-generating sub-network}
\label{subsec:constant network}
The thresholds $\varepsilon(\sigma)$ in Eqn. (\ref{eq:10}), which can be learned in a supervised learning manner, are crucial for the representation ability of sparse representation model-driven networks\cite{19}. Based on the so-called universal threshold theorem, the thresholds of soft shrinkage are proportional to the noise level, i.e., $\varepsilon_{i}=c_{i}\cdot\sigma$. Here $\varepsilon_{i}$ is the threshold for the $i$-th channel of the feature map, and $c_{i}$ is the corresponding proportional constant. Since image structures are varying in the spatial domain, the corresponding threshold should be spatially varying. In order to generate spatially varying thresholds, we elaborate the so-called CGNet which can generate proportional constants. The thresholds can be determined by the product of the proportional constants and the noise levels. Formally, the thresholds can be defined as
\begin{equation} 
	\bm{e}=[\varepsilon_{1},\varepsilon_{2},\cdots,\varepsilon_{i},\cdots]^{\rm T}=\bm{c}\otimes\bm{s}
	\label{eq:10_1}
\end{equation}
where $\otimes$ denotes the element-wise product, and $\bm{s}$ is the noise level map obtained by stretching the noise level $\sigma$. The size of $\bm{s}$ is the same as that of learned proportional constant vector $\bm{c}$. To avoid arbitrary values, the elements of $\bm{c}$ are limited to $[{c}_{min},{c}_{max}]$. By doing so, the thresholds are instance-optimal and spatial-variant. Since the image quality improves with each iteration, $\sigma$ should gradually decrease\cite{18}. In this paper, we introduce the learnable parameter $\tau \in(0,1)$, such that $\sigma_t=\tau\sigma_{t-1}$.\par
\noindent\textbf{The overall architecture of CGNet:} To provide a clearer understanding of CGNet, we present its overall architecture in Fig. \ref{fig4}. Specifically, the proposed CGNet consists of the shallow feature extraction module, the deep feature extraction module, the explicit prompt module, and the $3\times3$ convolutional layer, aiming to generate spatial-variant proportional constants to enhance the representation ability of the overall network. In shallow feature extraction, stacked convolutional layers perform filtering operations on the input to extract local features and structural information. Mathematically, the shallow feature extraction operator $\mathcal{F}_{shallow}(\cdot)$ can be expressed as follows:
\begin{equation} 
	\mathcal{F}_{shallow}(\cdot)=Conv(ReLU(Conv(\cdot)))
	\label{eq:1122}
\end{equation}
where $ReLU(\cdot)$ denotes the rectified linear unit activation layer and $Conv(\cdot)$ indicates the $3\times3$ convolutional layer. Building upon this, we introduce an explicit prompt module to encode specific image knowledge and sampling information, embedding it between shallow and deep feature extraction, thus enabling accurate discrimination of different sparse sampling settings. We define the prompt information obtained from the explicit prompt module as $\bm{p}$. At the $t$-th stage, the feature refinement process can be represented mathematically as follows:
\begin{equation} 
	\bm{q}_{t}=\mathcal{F}_{shallow}(\bm{W}\bm{z}_{t-1})\otimes\bm{p}+\bm{W}\bm{z}_{t-1}
	\label{eq:1123}
\end{equation}
where $\bm{Wz}_{t-1}$ is the coefficient vector and $\bm{q}_{t}$ denotes the feature map obtained through shallow feature extraction. Furthermore, in deep feature extraction, we employ the STB to compensate for the limitations of the receptive field imposed by convolutional operations, thereby capturing regional information on features and effectively extracting deep features. Subsequently, the SFB is employed to extract more comprehensive and detailed features. The process of extracting deep feature can be mathematically expressed as follows:
\begin{equation} 
	\mathcal{F}_{deep}(\cdot)=\mathcal{F}_{SFB}(\mathcal{F}_{STB}(\cdot))
	\label{eq:1124}
\end{equation}
where $\mathcal{F}_{deep}(\cdot)$ is the deep feature extraction operator, $\mathcal{F}_{STB}(\cdot)$ represents the procedure of the Swin Transformer block that captures global dependencies and long-range interactions in the feature space, and $\mathcal{F}_{SFB}(\cdot)$ denotes the spatial frequency block, which enhances the feature maps by emphasizing important spatial and frequency components to refine and strengthen the feature representation.﻿ Specifically, the fast fourier convolution for global information extraction in the frequency domain branch, and CNN-based residual modules for enhanced local feature representation in the spatial domain branch. Finally, the global skip connection combines shallow features with deep features to further enhance the model stability and the representation ability.\par 
To adaptively generate the thresholds and adapt to different sparse-view settings, a proportional constant vector is generated at each stage of the network. This constant vector plays a crucial role in balancing the contributions of different spatially varying features during the iterative process. The proportional constant-generating process at the $t$-th stage can be formulated as:
\begin{equation}
	\begin{aligned}
		\bm{c}_{t}=Conv(\mathcal{F}_{deep}(\mathcal{F}_{shallow}(\bm{W}\bm{z}_{t-1})\otimes\bm{p}+\bm{W}\bm{z}_{t-1}))+\bm{q}_{t})
		\label{CGNetEq2}
	\end{aligned}
\end{equation}
where $\bm{c}_{t}$ represents the corresponding output proportional constant map.\par
\noindent\textbf{Explicit prompt module:}
Prompt learning techniques have been widely employed in general image restoration as effective tools for various restoration tasks\cite{39}, each equipped with learnable prompts to interact with input images or latent features. Given the difficulty in acquiring a sufficient number of paired images in practical scenarios, learning prompts that can effectively discriminate between a large number of fine-grained sparse sampling is very challenging. To address this issue, we propose a more applicable explicit prompt module that learns discriminative information from input explicit prompts. Specifically, considering the differences in CT image distribution under different view numbers in the SVCT reconstruction task, we introduce a prompt module to effectively guide the generation of view-aware adaptive thresholds for different sparse view numbers. The essential difference in the sparse sampling configurations stems from the sampling distribution of projection data. Hence, we adopt a binary down-sampling matrix mask $\bm{M}\in\mathbb{R}^{{N_p}\times{N_b}}$ with 0 indicating the missing region, represented by the vector $\bm{m}$, naturally encoding the sampling information to construct explicit view prompts. We employ three convolutional layers to extract features related to sparse sampling from the down-sampling matrix, then apply full connection layer to convert the features into the suitable shape (1 $\times$ channel or dimension) for the corresponding module outputs. Formally, the explicit prompt module can be formulated by
\begin{equation} 
	\bm{p}=FC(F_{ext}(\bm{m}))
	\label{e22}
\end{equation}
where $\bm{p}$ indicates the prompt vector, $FC(\cdot)$ refers to the fully connected layer operation, and $F_{ext}(\cdot)$ denotes the CNN operation. We use convolution layers with a kernel size of $3\times3$ to extract feature information from the sampling settings and apply the ReLU activation function to enhance the ability of the network to model nonlinearities. By incorporating this prompt-based approach, LipNet can adaptively generate prompt features that carry discriminative information specific to each sampling setting, enabling universal SVCT reconstruction with a single model.
\subsection{Theoretical analysis of LipNet}
\label{Theoretical analysis of LipNet}
In this section, in order to demonstrate that LipNet satisfies the boundary property and Lipschitz constraint, we first present \textbf{Lemma 1} that proves the boundary property of LipNet, and then prove that LipNet satisfies the Lipschitz continuity. Based on \textbf{Lemma 1}, \textbf{Theorem 1} claims LipNet is a bounded network. Furthermore, \textbf{Theorem 2} claims the Lipschitz continuity of LipNet based on \textbf{Theorem 1}. The detailed proofs of \textbf{Lemma 1}, \textbf{Theorem 1}, and \textbf{Theorem 2} can be found in the supplementary material\footnote{\url{https://github.com/shibaoshun/PromptCT}}.\par
\noindent\textbf{Lemma 1:} For any input $\bm{x}\in\mathbb{R}^{n}$ and some universal constant $L=\tau_{0}c_{max}^{2}$ (Here, $c_{max}$ denotes the maximum element of proportional constant vector $\bm{c}$) independent of $M$, we can get
\begin{equation} 
	||\bm{W}^{\rm{T}}\bm{W}\bm{x}-\bm{W}^{\rm{T}}T_{\varepsilon(\sigma)}(\bm{W}\bm{x})||_2^2\leqslant{M}\sigma^{2}L
	\label{eq:12}
\end{equation}
where $M$ denotes the dimension of the threshold vector $\bm{e}$ and $\sigma$ indicates the input noise level.\par
\noindent\textbf{Theorem 1 (Boundary property of LipNet):} For any input $\bm{x}\in\mathbb{R}^{n}$, the proposed LipNet $\mathcal{D}_{\theta}(\cdot;\sigma)$ based on \textbf{Lemma 1} is bounded such that
\begin{equation} 
	||\bm{x}-\mathcal D_{\theta}(\bm{x};\sigma)||_2^2\leqslant{N}\sigma^{2}
	\label{eq:111}
\end{equation}
for some universal constant $N$ independent of $\sigma$.\par
\noindent\textbf{Theorem 2 (Lipschitz continuity of LipNet):} Based on \textbf{Theorem 1}, the proposed LipNet $\mathcal{D}_{\theta}(\cdot;\sigma)$ is $\upsilon$-Lipschitz continuous. This means that there exists a $\upsilon>0$ such that for all $\bm{x}_{1}, \bm{x}_{2}$, the following relationship holds
\begin{equation} 
	||\mathcal D_{\theta}(\bm{x}_{1};\sigma)-\mathcal D_{\theta}(\bm{x}_{2};\sigma)||_2^2\leqslant\upsilon||\bm{x}_{1}-\bm{x}_{2}||_2^2.
	\label{eq:31}
\end{equation}
\section{The proposed PromptCT for SVCT reconstruction}
\label{sec:PromptCT}
\subsection{Problem formulation and deep unfolding network}
\label{subsec:Problem Formulation}
In fan-beam SVCT system configuration, sparse-view projection data $\bm{Y}\in\mathbb{R}^{{N_p}\times{N_b}}$ (a.k.a raw projection or sinogram), obtained through uniform sampling via a $360$-degree rotation of the detector, where ${N_p}$ and ${N_b}$ respectively represent the number of sparse-view projections and detector bins. The SVCT reconstruction aims to recover the underlying CT image from the observed incomplete projection data, for which we can formulate the corresponding minimization problem as follows:
\begin{equation} 
	\mathop{\min}\limits_{\bm{x}}\frac{1}{2}||\bm{y}-\mathcal{P}\bm{x}||_2^2+\lambda{R}(\bm{x})
	\label{eq:1}
\end{equation}
where $\mathcal{P}$ is the forward projection operator, specifically the Radon transform under sparse-view conditions. We define the under-sampling data as $\bm{y}=vec(\bm{Y})\in\mathbb{R}^{{N_p}{N_b}}$ using the vector operator. The first term is known as the data-fidelity term, which encourages consistency between the recovered CT image and the projection data. The second term represents the regularization term, imposing desired properties on the image to be reconstructed. The parameter $\lambda$ is a regularization parameter that pursues the trade-off between these two terms.\par
The proximal gradient descent algorithm applied to the optimization problem in Eqn. (\ref{eq:1}) yields the following iterations
\begin{numcases}{}
	\bm{x}^{(k+0.5)} =\bm{x}^{(k)}+\eta\mathcal{P}^{\rm{T}}(\bm{y}-\mathcal{P}\bm{x}^{(k)}) \label{eq:2_2} \\	
	\bm{x}^{(k+1)} = prox_{\eta R}(\bm{x}^{(k+0.5)})
	\label{eq:2_3}
\end{numcases}
where $\eta>0$ is the step size and $prox_{\eta R}(\cdot)$ is a proximal operator related to the prior form ${R}(\cdot)$ about $\bm{x}$. The fundamental idea behind the deep unfolding approach is to fix a maximum number of iterations denoted as $K$, and declare $\bm{x}^{(K)}$ as our estimate for $\bm{x}$. We refer to Eqn. (\ref{eq:2_2}) as the image update module (IUM) and consider replacing $prox_{\eta R}(\cdot)$ with a trainable prior network $\mathcal{D}_{\theta}(\cdot;\sigma)$, which gives the iteration map
\begin{equation} 
	\bm{x}^{(k+1)} = \mathcal{D}_{\theta}(\bm{x}^{(k+0.5)};\sigma).
	\label{eq:2_1}
\end{equation}\par
\begin{figure}[!h]
	\centerline{\includegraphics[width=\columnwidth]{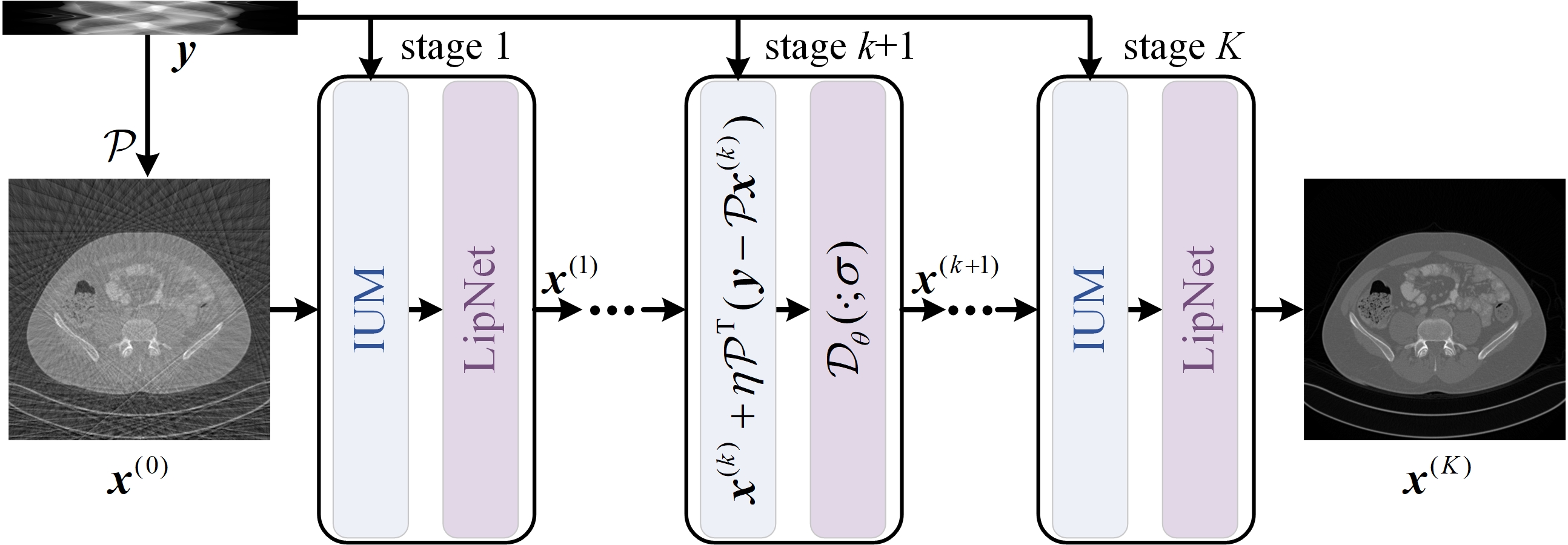}}
	\caption{The proposed deep unfolding proximal gradient descent network architecture (i.e., PromptCT) consists of the IUM and the prompting Lipschitz-constrained network $\mathcal{D}_{\theta}(\cdot;\sigma)$ (i.e., LipNet).}
	\vspace{-0.4cm}
	\label{fig5}
\end{figure}

The prior network $\mathcal{D}_{\theta}(\cdot;\sigma)$ defined in Eqn. (\ref{eq:2_1}) is crucial for enhancing reconstruction quality. Therefore, we employ the well-designed LipNet as the prior network to improve the reconstruction performance through strong representation ability. Figure. \ref{fig5} shows the overall network architecture of the proposed deep unfolding proximal gradient descent network for the SVCT reconstruction task. Specifically, PromptCT consists of $K$ stages corresponding to $K$ iterations of the iterative algorithm, and each stage contains IUM and LipNet.
\subsection{Loss Functions}
\label{subsec:Loss Functions}
As mentioned earlier, we design a prompt-based deep unfolding network for the SVCT reconstruction task. For the training process of PromptCT, in order to remove artifacts while preserving the global structures of the output CT image, we supervise the restored CT image $\bm{x}^{(k)}$ updated at each stage using the ${l}_1$ loss and the ${l}_2$ loss. The total loss function of training PromptCT can be formulated as
\begin{equation}
	\begin{aligned}	
		\mathcal{L}=&\sum_{k=0}^{K}\Big\{\omega_1||\bm{x}^{(k)}-\bm{x}_{gt}||_1+\omega_2||\bm{x}^{(k)}-\bm{x}_{gt}||_2^2\Big\}\\
		&+||\bm{x}-\bm{x}_{gt}||_2^2
		\label{lossEq1}
	\end{aligned}
\end{equation}
where $\bm{x}_{gt}$ is the ground truth CT image. Meanwhile, $\omega_1$ and $\omega_2$ are hyper-parameters to balance the weights of different loss items. Heuristically, we set $\omega_1=0.1$ and $\omega_2=0.1$.
\begin{table*}[!t]
	%	\vspace{-0.3cm}
	\centering
	\caption{Quantitative evaluations for different SVCT methods on the AAPM dataset under various sparse-view conditions. We report the average PSNR (dB)$\color{red}{\uparrow}$/ SSIM$\color{red}{\uparrow}$/ RMSE$\color{red}{\downarrow}$ values of the testing dataset for each case. The best results are highlighted in {\color{red}red} and the second-best results are highlighted in {\color{blue}blue}.}
	\footnotesize{\tabcolsep=10pt}
	%	\vspace{-0.1cm}  %调整图片与上文的垂直距
	\resizebox{\linewidth}{!}{
		\begin{tabular}{cccccccccccccccc}
			\toprule[1.3pt]
			\multirow{2}{*}{\bf Method}&
			\multicolumn{3}{c}{\bf 60}&\multicolumn{3}{c}{\bf 90} &\multicolumn{3}{c}{\bf 120} &\multicolumn{3}{c}{\bf 180} &\multicolumn{3}{c}{\bf average}
			\cr\cmidrule(r){2-4} \cmidrule(r){5-7} \cmidrule(r){8-10} \cmidrule(r){11-13} \cmidrule(r){14-16}
			&PSNR&SSIM&RMSE&PSNR&SSIM&RMSE&PSNR&SSIM&RMSE&PSNR&SSIM&RMSE&PSNR&SSIM&RMSE\\\midrule
			\multicolumn{16}{c}{single-view models}\\\midrule		
			FBP        &13.06&0.3884&0.2242&14.47&0.5087&0.1906&16.16&0.6160&0.1570&20.99&0.7746&0.0900&16.17&0.5719&0.1655\\
			RED-CNN\cite{7}    &36.09&0.9129&0.0159&38.28&0.9309&0.0123&39.78&0.9429&0.0103&41.63&0.9516&0.0083&38.95&0.9346&0.0117\\
			FBPConvNet\cite{8} &34.74&0.9036&0.0188&37.23&0.9249&0.0140&38.99&0.9374&0.0114&41.35&0.9540&0.0086&38.08&0.9300&0.0132\\
			DD-Net\cite{24}    &35.50&0.8762&0.0170&37.28&0.8668&0.0138&38.46&0.8927&0.0120&40.29&0.9133&0.0097&37.88&0.8873&0.0131\\
			CNNMAR\cite{10}    &36.11&0.9155&0.0159&38.42&0.9358&0.0122&39.77&0.9458&0.0104&41.97&{\color{blue}0.9582}&0.0080&39.07&0.9388&0.0116\\
			MSANet\cite{45}    &36.20&0.9170&0.0157&38.36&0.9331&0.0122&39.91&0.9445&0.0102&41.93&0.9580&0.0081&39.10&0.9382&0.0116\\
			DuDoTrans\cite{9}  &35.60&0.9049&0.0168&38.11&0.9274&0.0125&39.75&0.9405&0.0104&{\color{blue}42.03}&0.9571&{\color{blue}0.0080}&38.87&0.9325&0.0119\\
			FreeSeed\cite{29}  &35.68&0.9005&0.0167&37.86&0.9257&0.0129&39.25&0.9376&0.0110&41.25&0.9524&0.0087&38.51&0.9291&0.0123\\
			LipCT (ours)&{\color{blue}36.38}&{\color{blue}0.9184}&{\color{blue}0.0154}&{\color{blue}38.84}&{\color{blue}0.9374}&{\color{blue}0.0115}&{\color{blue}40.10}&{\color{red}0.9466}&{\color{blue}0.0100}&{\color{red}42.16}&{\color{red}0.9598}&{\color{red}0.0078}&{\color{blue}39.37}&{\color{blue}0.9406}&{\color{blue}0.0112}\\	
			\midrule\multicolumn{16}{c}{multi-view models}\\ \midrule
			MLipCT (ours)&36.16&0.9167&0.0158&38.13&0.9309&0.0126&39.38&0.9399&0.0109&40.80&0.9528&0.0092&38.62&0.9351&0.0121\\
			{PromptCT} (ours)&{\color{red}36.95}&{\color{red}0.9224}&{\color{red}0.0144}&{\color{red}38.95}&{\color{red}0.9375}&{\color{red}0.0114}&{\color{red}40.16}&{\color{blue}0.9464}&{\color{red}0.0099}&41.91&0.9578&0.0081&{\color{red}39.49}&{\color{red}0.9410}&{\color{red}0.0110}\\
			\bottomrule[1.3pt]
	\end{tabular}}
	\vspace{-0.3cm}
	\label{tab2}
\end{table*}
\subsection{Theoretical analysis}
\label{subsec:Theoretical analysis}
Classical fixed-point theory indicates that the iterations converge to a unique fixed-point if the iteration map $f_{\theta}(\cdot;\bm{y})$ is contractive\cite{DEQ}, i.e., there exists a constant $0 \leq c < 1$ such that $||f_{\theta}(\bm{x}_1;\bm{y})- f_{\theta}(\bm{x}_2;\bm{y})||_2^2 \leq c||\bm{x}_1-\bm{x}_2||_2^2$ for all $\bm{x}_1, \bm{x}_2 \in \mathbb{R}^d$. Based on this observation, we derive the following convergence result for PromptCT:\par
\noindent\textbf{Theorem 3 (Convergence of PromptCT):} According to \textbf{Theorem 2}, the network $\mathcal D_{\theta}(\cdot;\sigma)$ is $\upsilon$-Lipschitz. Assume that $\epsilon=\delta_{min}(\mathcal{P}^{\rm{T}}\mathcal{P})$ and $\zeta=\delta_{max}(\mathcal{P}^{\rm{T}}\mathcal{P})$, where $\delta_{min}(\cdot)$ and $\delta_{max}(\cdot)$ denote the maximum and minimum eigenvalue, respectively. Then the iteration map of the proposed PromptCT, i.e., $f_{\theta}(\cdot; \bm{y})$, is contractive if the step size parameter $\eta$ satisfies
\begin{equation} 
	\frac{1}{\epsilon}-\frac{1}{\upsilon\epsilon}<\eta<\frac{1}{\zeta}+\frac{1}{\upsilon\zeta}
	\label{eq:34}
\end{equation}
Such an $\upsilon$ exists if $\upsilon<(\epsilon+\zeta)/(\zeta-\epsilon)$.\par
\noindent\textit{Proof:} See the supplementary material\footnote{\url{https://github.com/shibaoshun/PromptCT}}.\par
\section{Experiments}
\label{sec:Experiments}
\subsection{Experimental setup}
\label{subsec:Experimental setup}
\subsubsection{Datasets}
\label{subsubsec:Datasets}
We perform experiments on two publicly available real-world CT image datasets: the 2016 NIH-AAPM-Mayo Clinic Low Dose CT Grand Challenge dataset \cite{9e8782ab41f14bb398bd595424a08194} and the DeepLesion dataset \cite{Yan2018DeepLesionAM}. To evaluate the feasibility of our method in practical applications, experiments are conducted on real objects, including the equivalent water-bone phantom \cite{Xue2019ImageRF} and the pork with bone slices. The experimental setup is detailed below. The phantom is scanned using the industrial CT system in our laboratory. The X-ray source is HAMAMATSU-L12161-07, and the detector used is the EIGER2 1MW R-DECTRIS. The voltage and current settings for the X-ray source are 100 kVp and 500 $\mu$A, respectively. The detector integration time is set to 2 seconds. The distance from the X-ray source to the rotation center is 220.798 mm, and the distance from the rotation center to the detector is 197.247 mm. The detector has a unit size of 0.3 mm with 512 units in total. The sampling angular range spans 0 to $360^{\circ}$, with 60 angular sampling points. For simplicity, the experiments are restricted to the 2D fan-beam case. There should be no difficulty to extend it to the 3D cone-beam case. To validate that our method can be extended to 3D scenarios, we select the vertebrae localization and identification dataset from SpineWeb \cite{10.1007/978-3-642-40763-5_33}, which includes 14 testing volumes labeled with multi-bone, i.e., sacrum, left hip, right hip, and lumbar spine. The clinical images are resized and processed using the same protocol to the synthesized data. We conduct experiments on the SpineWeb dataset and generate axial, coronal, and sagittal CT images for evaluation.\par
The normal-dose Mayo data are acquired using a 120 kVp and 235 effective mAs (500 mA/0.47 s) protocol, scanning from the chest to the abdomen. We select 1000 images for training and 500 images for testing, with each image resized to $512\times512$. The experiments are performed in a fan-beam X-ray source scanning setup with 800 detector elements. We generate a fully sampled sinogram by acquiring 360 projection views at regular intervals between 0 and 360 degrees. We obtain sparse-view projection data by uniformly sampling 60, 90, 120, and 180 views from the full-view sinogram. In the numerical experiments, the simulated sinogram is contaminated by Poisson noise with an intensity of $5\times10^{6}$ and electronic system noise (with standard deviation $5\%$). The electronic noise level is regarded to follow a zero-mean normal distribution and is assumed to be stable for a commercial CT scanner. For the DeepLesion dataset, we randomly choose 1000 images as the training set and another 200 images as the test set. Concretely, we use a 120 kVp polyenergetic X-ray source to simulate the equiangular fan-beam projection geometry and the incident X-ray contains $2\times10^{7}$ photons. We specify that the full-sampling sinograms are generated by uniformly spaced 360 projection views between 0 and 360 degrees, and the number of detector elements is 641. The corresponding artifact-affected CT images are reconstructed from the sparse-view sinograms by FBP, and all CT images are resized to $416\times416$. To simulate the photon noise numerically, we add mixed noise that is by default composed of $0.2\%$ Gaussian noise and Poisson noise with an intensity of $2\times10^{7}$. To further validate the effectiveness of our approach, we reconstructed images with a resolution of $512\times512$ during testing to further evaluate its generalization capability in handling images with a larger number of pixel values.\par
\begin{figure*}[!t]
	\centering
	\includegraphics[width=1\linewidth]{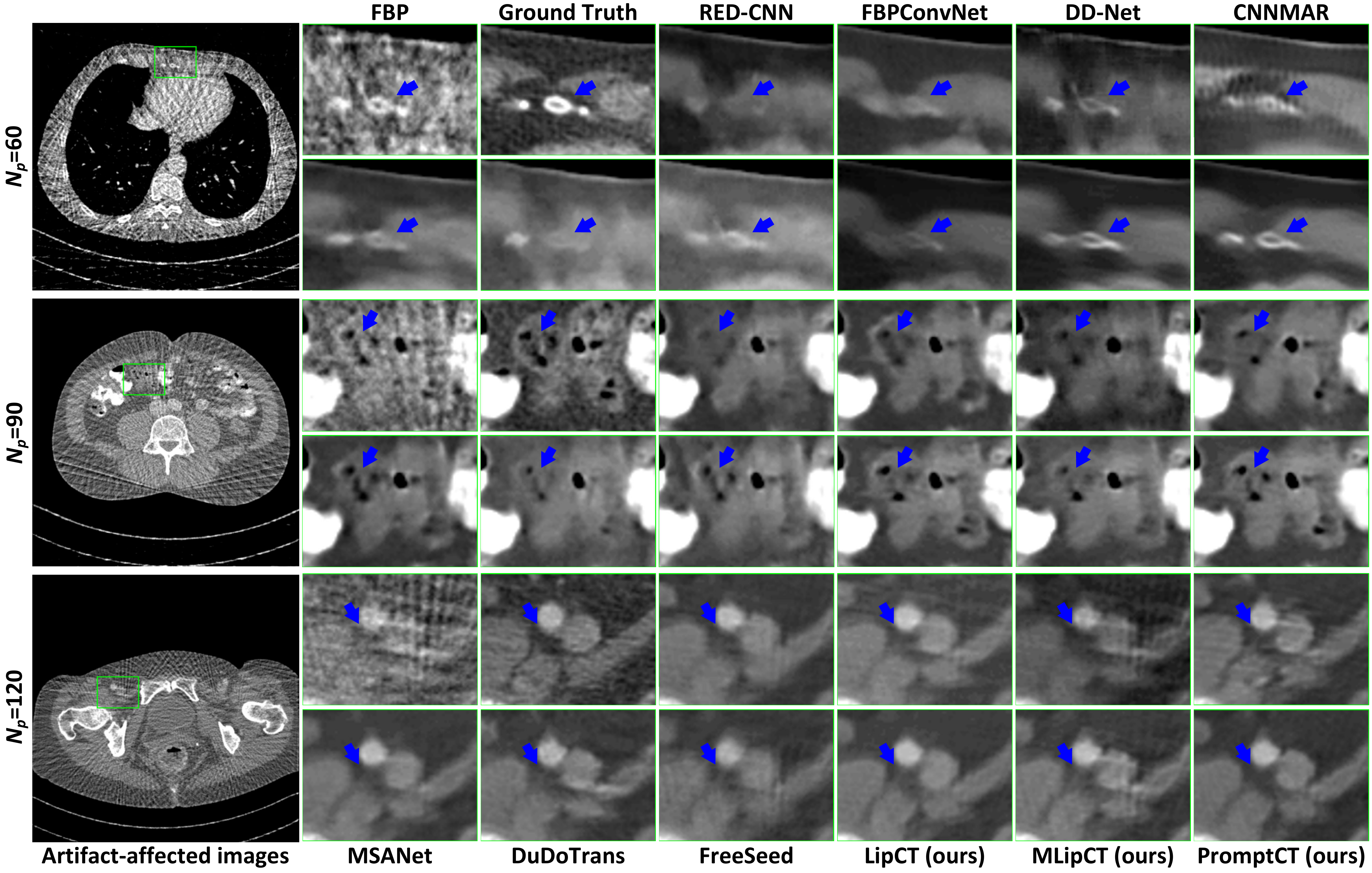}
	\caption{Visual comparison of SVCT reconstruction methods on AAPM dataset under different sparse-view settings. The display window is [-175, 500] HU. Regions of interest are zoomed in for better viewing. Blue arrows indicate key areas, such as the sternum, intestinal tissues, and tissues near the femoral artery, which highlight differences in reconstruction quality among the methods. For single-view models, our LipCT outperforms other benchmark methods. For multi-view models, our PromptCT method, which incorporates explicit prompts, achieves better performance than that of MLipCT.}
%	\vspace{-0.4cm}
	\label{fig6}
\end{figure*}
\begin{table*}[!h]
	%	\vspace{-0.3cm}
	\centering
	\caption{Storage requirement for different models. In comparison regarding the storage costs (MB) of models for four sparse views, the multiple-in-one model demonstrates better storage efficiency.}
	\footnotesize{\tabcolsep=10pt}
	%	\vspace{-0.1cm}  %调整图片与上文的垂直距
	\resizebox{\linewidth}{!}{
		\begin{tabular}{ccccccccccc}
			\toprule[1.3pt]
			{\bf Method}&{\bf RED-CNN}&{\bf FBPConvNet}&{\bf DD-Net}&{\bf CNNMAR}&{\bf MSANet}&{\bf DuDoTrans}&{\bf FreeSeed}&{\bf LipCT (ours)}&{\bf MLipCT (ours)}&{\bf {PromptCT} (ours)}\\
			\midrule
			Storage	&84.7&1582.8&26.8&125.1&396.1&1054.6&357.7&37.9&9.5 &12.5\\
			\bottomrule[1.3pt]
	\end{tabular}}
	\vspace{-0.3cm}
	\label{tab5}
\end{table*}
\subsubsection{Training Details}
\label{subsubsec:Training Details}
Our method is implemented using the PyTorch framework, and we use the differentiable operation $\mathcal{P}$ and $\mathcal{P}^{\rm{T}}$ operation in ODL\footnote{\url{https://github.com/odlgroup/odl}} library on the NVIDIA RTX 3090Ti GPU. The ADAM algorithm is employed as the optimizer with $(\beta_{1}, \beta_{2}) = (0.5, 0.999)$ whose initial learning rate is $1\times10^{-4}$. The number of training epochs is 100 with a batch size of 1. Heuristically, we manually set $\eta=2\times10^{-4}$, $\alpha=0.1$, $\beta=9$, and $\tau=0.8$ in our experiments. For training, we consider sparse view numbers of 60, 90, 120, and 180. Specifically, we train single-view models for each view separately. Moreover, the multi-view models are trained using a hybrid-view learning strategy, with the training data increasing from an initial training set of 1000 to 4000. It is worth noting that we did not add any additional original projection data to ensure fairness in experimental comparisons.
\subsection{Compared with previous SVCT algorithms}
\label{subsec:Compared with previous SVCT algorithms}
To validate the effectiveness and generalization performance of our proposed method, we perform experiments under two different experimental setups: single-view models with separate training and multi-view models with mixed training. Specifically, we divide the proposed approaches into three versions for experiments: (\textit{\romannumeral1}) separately trained single-view models, denoted as LipCT, where CGNet excludes the prompt module; (\textit{\romannumeral2}) mixed trained multi-view models, denoted as MLipCT, where CGNet excludes the prompt module; and (\textit{\romannumeral3}) mixed trained multi-view models, where CGNet includes the prompt module, denoted as PromptCT. It is worth noting that existing methods need to train the models separately for different sparse view configurations, whereas our proposed MLipCT and PromptCT only need to be trained once at the case of multiple views.
\subsubsection{Quantitative results}
\label{subsubsec:Quantitative results}
We compare the proposed PromptCT with the current DL-based SVCT reconstruction methods including RED-CNN\cite{7}, FBPConvNet\cite{8}, DD-Net\cite{24}, CNNMAR\cite{10}, MSANet\cite{45}, DuDoTrans\cite{9}, and FreeSeed\cite{29}. We employ the peak signal-to-noise ratio (PSNR), structured similarity index (SSIM), and root mean square error (RMSE) for perceived visual quality assessment. Table \ref{tab2} reports the quantitative comparison of different SVCT reconstruction methods. Specifically, we validated the effectiveness of the proposed method at multiple sampling views ($N_{p}=60, 90, 120, 180$). From the results of the single-view models, it can be seen that our proposed deep unfolding approach (i.e., LipCT) outperforms the previous methods in most sampling settings and achieves the best average value, attributed to the superior representation ability of the proposed prior network. We observe that the DL-based dual-domain methods (i.e., CNNMAR) exhibit significant superiority over the image-domain methods due to the richness of information provided by the sinogram. However, as the number of sparse views decreases, the available information provided by the sinogram also decreases and the advantage of the dual-domain approach is no longer significant.\par
Single-view models require complicated training processes for each view as well as extensive storage costs, while the generalizability of single-view models is limited. Therefore, we employ the mixed training strategy to unlock multiple sparse-view settings using a single model. Consistent with our proposed single-view model architecture, we try to use mixed training instead of single-view training. The results indicate that without using prompts, training with multi-view data concurrently may adversely affect the learning process and lead to suboptimal performance. In contrast, our proposed PromptCT achieves higher PSNR/SSIM values and lower RMSE values compared to the single-view models, suggesting that the non-trivial transferability of PromptCT stems from the prompt-based model rather than the training strategy.\par
\subsubsection{Qualitative results}
\label{subsubsec:Qualitative results}
Figure \ref{fig6} shows representative sliced results for different methods in different sparse sampling views, where the region of interest is zoomed in to aid visualization. As shown by the blue arrows in Fig. \ref{fig6}, previous DL-based methods are successful in removing a large number of streak artifacts in some settings, but they introduce secondary artifacts, especially when the number of sparse views is small. These issues are especially pronounced in the regions near the sternum, intestinal tissues, and tissues surrounding the femoral artery, where clear boundaries and fine textures are critical for diagnostic accuracy.\par
In contrast, our proposed LipCT effectively removes most streak artifacts while preserving the tissue structures of CT images. However, there is still room for improvement when the number of sampling angles is significantly reduced. For multi-view models, directly training LipCT with multi-view mixed data (i.e., MLipCT) results in suboptimal reconstruction performance, with noticeable degradation in reconstruction accuracy. However, our proposed PromptCT, by introducing explicit prompts, further improves the reconstruction quality. It not only eliminates artifacts more effectively but also preserves finer tissue details and structure integrity, even under highly sparse settings such as $N_{p}=60$. For instance, in the blue-arrow-indicated regions, PromptCT demonstrates superior performance by reconstructing the contours of the sternum with high precision and preserving the subtle textures of intestinal tissues, which are often blurred by other methods. These regions clearly showcase the ability of PromptCT to deliver sharper and more accurate reconstructions compared to existing approaches, highlighting its effectiveness in addressing SVCT challenges.\par
\begin{table}[!h]
	\centering
	\caption{Quantitative evaluation [PSNR (dB)$\color{red}{\uparrow}$/ SSIM$\color{red}{\uparrow}$/ RMSE$\color{red}{\downarrow}$] of different SVCT methods on the DeepLesion dataset at the sparse view number of 60. The best results are highlighted in {\color{red}red} and the second-best results are highlighted in {\color{blue}blue}.}
	\footnotesize{\tabcolsep=10pt}
	\resizebox{0.7\linewidth}{!}{
		\begin{tabular}{cccc}
			\toprule[1.3pt]
			\textbf{Method}   &\textbf{PSNR} &\textbf{SSIM} &\textbf{RMSE}\\
			\midrule
			\multicolumn{4}{c}{single-view models}\\\midrule		
			FBP        &13.33&0.4466&0.2202\\
			RED-CNN\cite{7}    &37.63&0.9426&0.0134\\
			FBPConvNet\cite{8} &37.63&0.9518&0.0134\\
			DD-Net\cite{24}    &37.90&0.9463&0.0130\\
			CNNMAR\cite{10}    &37.69&0.9566&0.0133\\
			MSANet\cite{45}    &36.94&0.9405&0.0145\\
			DuDoTrans\cite{9}  &37.87&0.9383&0.0130\\
			FreeSeed\cite{29}  &37.71&0.9397&0.0132\\
			LipCT (ours)  &{\color{blue}39.11}&{\color{red}0.9581}&{\color{blue}0.0113}\\ \midrule
			\multicolumn{4}{c}{multi-view models}\\\midrule	
			MLipCT (ours) &35.38&0.9157&0.0173 \\
			{PromptCT} (ours) &{\color{red}39.17}&{\color{blue}0.9576}&{\color{red}0.0112}\\
			\bottomrule[1.3pt]
	\end{tabular}}
%	\vspace{-0.3cm}
	\label{tab4}
\end{table}
\begin{figure}[!h]
	\centering
	\includegraphics[width=1\linewidth]{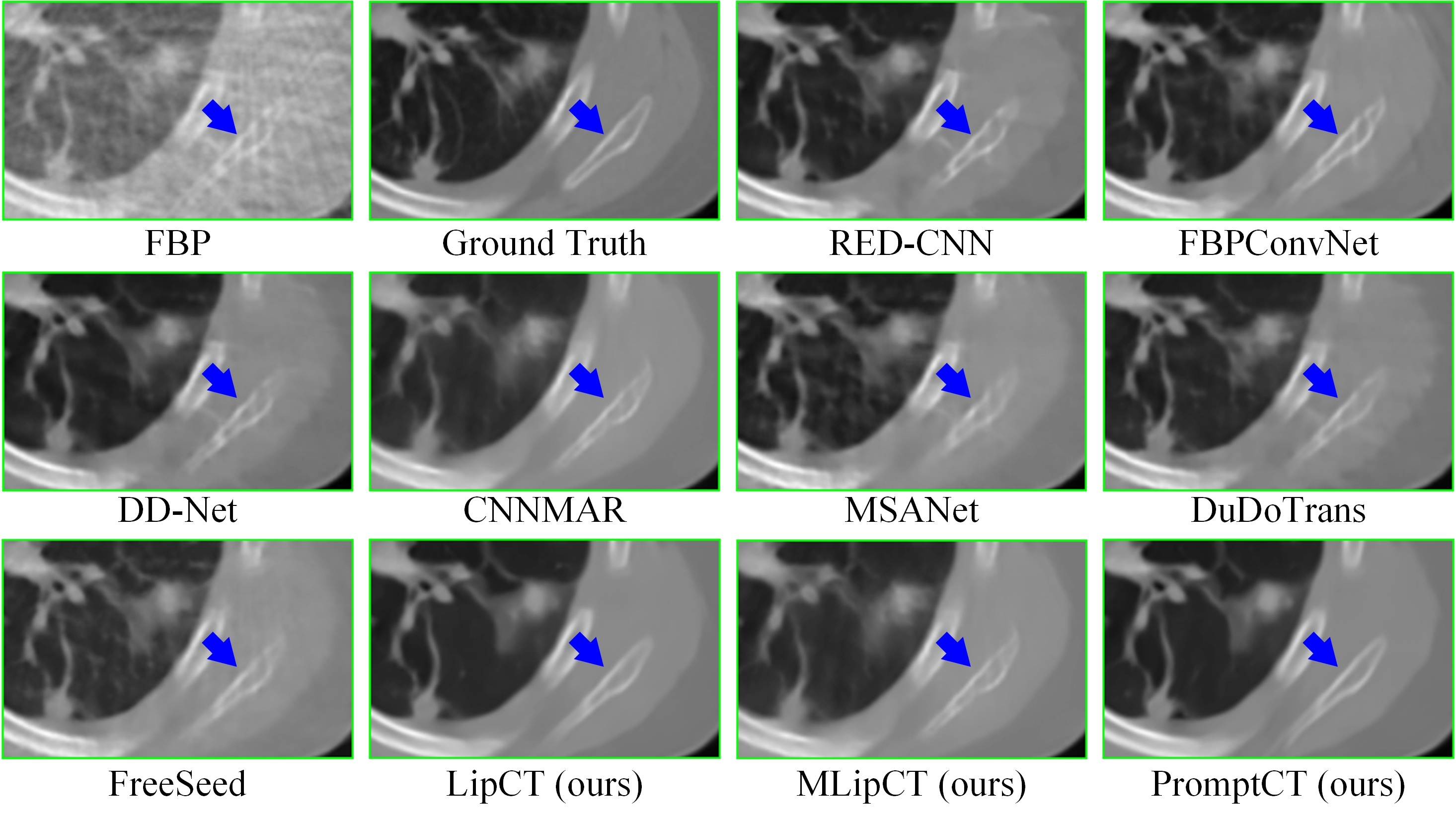}
	\caption{Visual comparison of different SVCT methods on the DeepLesion dataset at the sparse view number of 60. Regions of interest are zoomed in for better viewing. Blue arrows indicate the bone structure and highlight the differences in reconstruction quality among the methods. For single-view models, our LipCT outperforms other benchmark methods. For multi-view models, our PromptCT method, which incorporates explicit prompts, achieves better performance than that of MLipCT.}
	\vspace{-0.3cm}
	\label{fig13}
\end{figure}
\begin{figure*}[!h]
	\centering
	\includegraphics[width=1\linewidth]{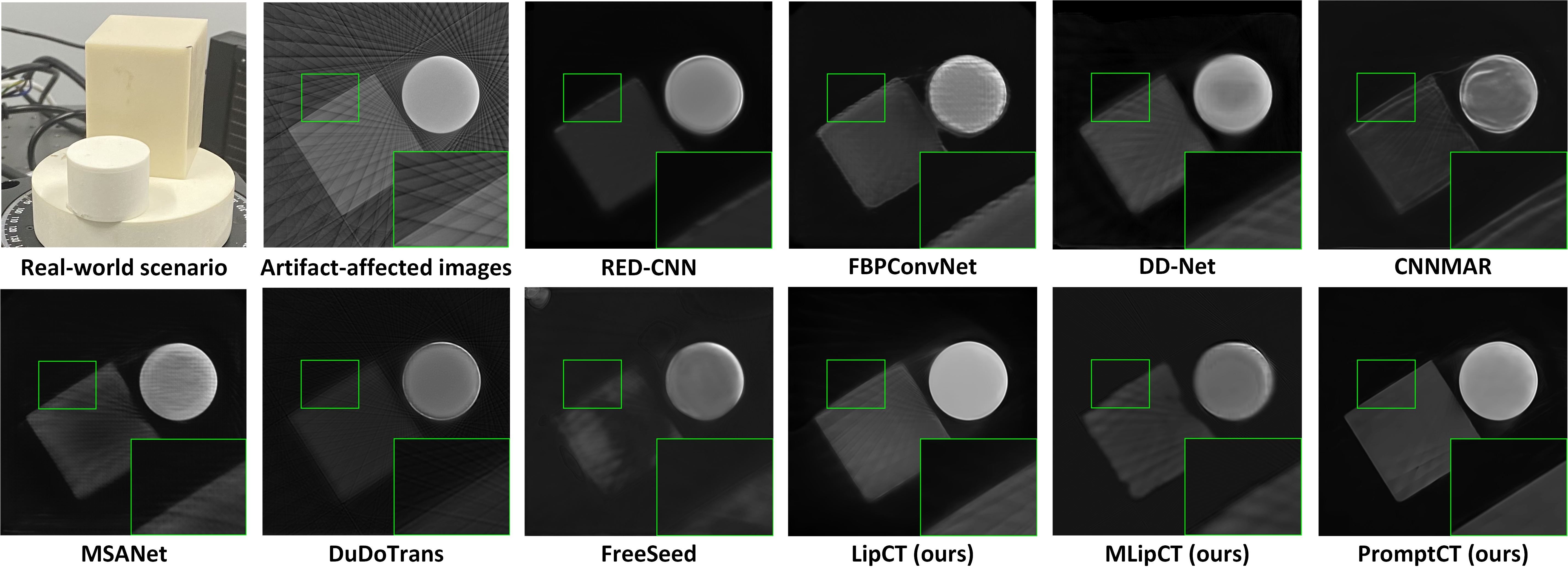}
	\caption{Visual comparison of SVCT reconstruction methods on the equivalent water-bone phantom experiment at the sparse view number of 60. Regions of interest are zoomed in for better viewing. For single-view models, our LipCT outperforms other benchmark methods. For multi-view models, our PromptCT method, which incorporates explicit prompts, achieves better performance than that of MLipCT.}
%	\vspace{-0.4cm}
	\label{fig10}
\end{figure*}
\begin{figure*}[!h]
	\centering
	\includegraphics[width=1\linewidth]{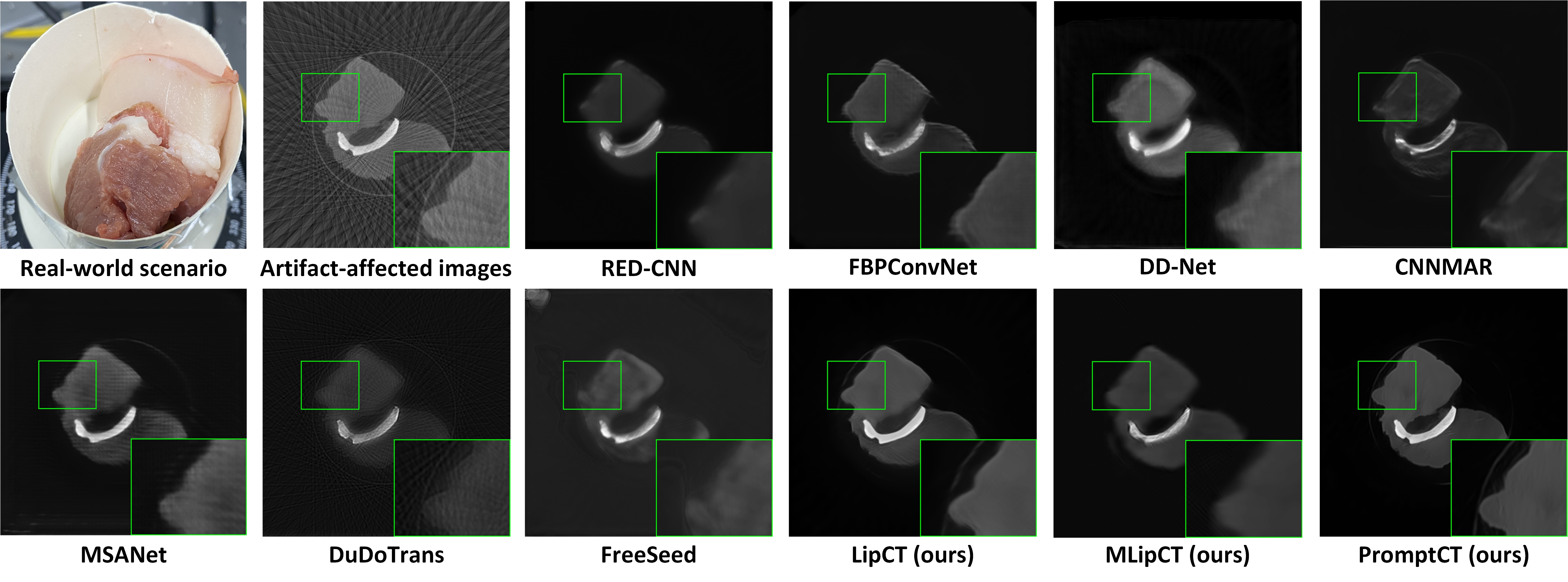}
	\caption{Visual comparison of SVCT reconstruction methods on the pork with bone slices experiment at the sparse view number of 60. Regions of interest are zoomed in for better viewing. For single-view models, our LipCT outperforms other benchmark methods. For multi-view models, our PromptCT method, which incorporates explicit prompts, achieves better performance than that of MLipCT.}
	\vspace{-0.4cm}
	\label{fig11}
\end{figure*}
\subsubsection{Storage costs of models}
\label{subsubsec:Storage costs of models}
In addition, we compute the storage costs of the models to comprehensively evaluate the performance of the algorithm. We provide the comparison of storage requirements in Tab. \ref{tab5}. The single-view models create a large storage requirement since it requires saving a model for each view in Tab. \ref{tab2}. In contrast, only one model of the multi-view models needs to be stored. Moreover, the proposed strategy achieves a good trade-off between performance improvement and saving storage costs.
\subsubsection{Comparison on DeepLesion dataset}
\label{subsubsec:Comparison on DeepLesion dataset}
To further validate the generalizability of our method across different datasets, we conducted training and evaluation on the DeepLesion dataset. Notably, we trained these models using commonly used $416\times416$ images and reconstructed larger $512\times512$ images during testing to further evaluate the ability of these methods to handle high-resolution images. Table \ref{tab4} presents a performance comparison of various SVCT reconstruction methods under the sparse-view condition with 60 sampling views. Among the single-view models, DL-based methods demonstrate significant improvements in reconstruction quality compared to traditional methods under this condition. However, their performance metrics are still suboptimal. Our proposed LipCT outperforms other single-view methods due to its ability to effectively integrate structural and contextual information through its tailored architecture. Furthermore, the multi-view model PromptCT, with explicit prompting, demonstrates superior performance compared to MLipCT, which does not incorporate prompts, under the sparse-view condition. This suggests that the remarkable transferability of PromptCT stems from its prompt-based design rather than the training strategy. As shown in Fig. \ref{fig13}, PromptCT achieves the best overall performance, owing to the incorporation of explicit prompts, which enhance the adaptability of the model to different sparse-view scenarios and enable it to preserve intricate details even in challenging regions. This underscores the effectiveness of our approach in improving generalization and reconstruction quality under sparse-view conditions.
%Additional quantitative results for other sparse-view angles, along with visual results and analyses, are provided in the supplementary material\footnote{\url{https://github.com/shibaoshun/PromptCT}}.}
\subsubsection{Comparison on real experimental data}
\label{subsubsec:Comparison on real experimental data}
To validate the generalizability of our proposed methods, we conduct a comparison of SVCT reconstruction methods on real experimental data, analyzing both the equivalent water-bone phantom experiment and the pork with bone slices experiment. For the equivalent water-bone phantom experiment, we use an artificial synthetic material with attenuation coefficients that approximate those of water and bone, where the cylinder represents the equivalent bone, and the cube represents the equivalent water. Figure \ref{fig10} shows the reconstruction results for the equivalent water and bone geometry phantom experiment. Data-driven methods, such as RED-CNN and FBPConvNet, reduce some artifacts but tend to excessively smooth the images, resulting in blurred details and loss of structures. Other DL-based methods, such as FreeSeed, still face challenges in preserving fine details, particularly at the edges of water-like tissue structures. In contrast, LipCT outperforms other single-view methods by effectively removing artifacts while preserving structural integrity, especially around water-tissue regions. For multi-view models, MLipCT struggles with artifact removal due to the lack of prompt information. In comparison, PromptCT demonstrates superior performance in handling equivalent water and bone structures, excelling in recovering edge information with great precision. For the pork with bone slice experiment, we scan real pork and insert a thin slice of bone within the pork. As shown in Fig. \ref{fig11}, comparisons of the reconstruction results on pork and bone slices further validate the effectiveness of our approach. LipCT performs better than other single-view methods, managing to preserve the complex boundaries between soft tissue and bone. PromptCT again leads in performance, delivering the most accurate reconstruction, particularly in the detailed regions of both bone and pork slices, demonstrating its effectiveness in preserving subtle tissue structures in challenging conditions.\par
\subsection{Ablation studies}
\label{subsec:Ablation studies}
\subsubsection{The effect of different network architectures in LipNet}
We conduct a series of ablation experiments to analyze the effectiveness of network architectures in the proposed backbone of LipNet at $N_{p}=120$. In the ablation experiments conducted for proposed LipNet, all other hyperparameters were kept constant, while maintaining the network structure as determined in its final form. The configurations are in five groups: (\textit{\romannumeral1}) Net-1: replacing LipNet with CNNs; (\textit{\romannumeral2}) Net-2: replacing our designed CGNet with a learnable parameter; (\textit{\romannumeral3}) Net-3: replacing LipNet with a tight frame where $\bm W$ is fixed; (\textit{\romannumeral4}) Net-4: replacing LipNet with a tight frame network containing tight constraint; (\textit{\romannumeral5}) Net-5: our proposed LipNet, where $\bm W$ is adaptively learned without any tight constraints.\par
\begin{figure}[!h]
	\centerline{\includegraphics[width=\linewidth]{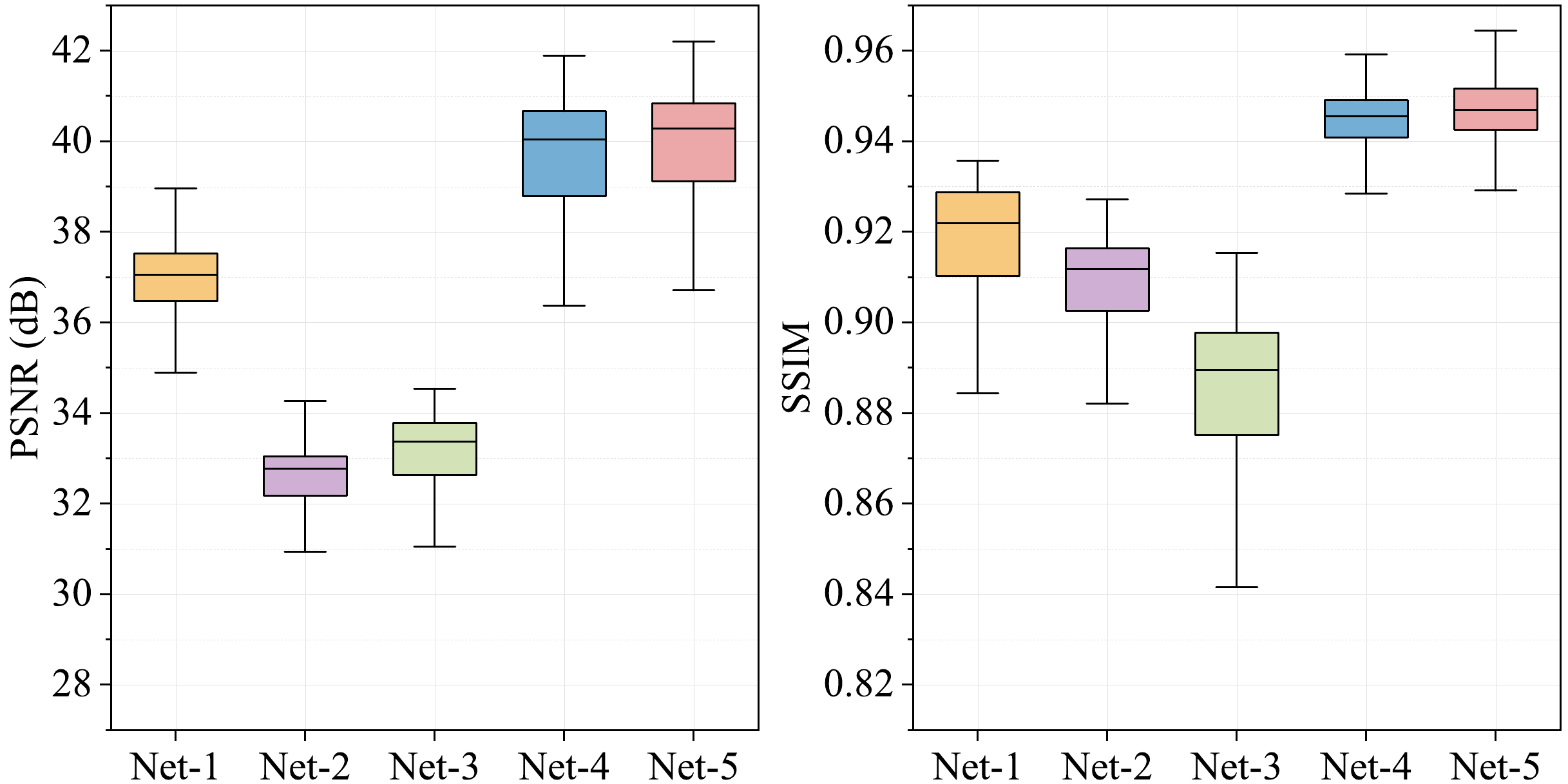}}
	\caption{Boxplots for PSNR(dB)/SSIM of different configurations at the sparse view number of 120. The left boxplot illustrates the PSNR values on the AAPM test dataset. The right boxplot displays the distribution of SSIM values.}
	\vspace{-0.4cm}
	\label{fig7}
\end{figure}
Figure \ref{fig7} presents the corresponding boxplots for PSNR/SSIM, where Net-5 is our baseline. When LipNet is replaced with simple CNNs in Net-1, the performance significantly degrades due to the insufficient representation ability provided by CNNs. Similarly, substituting CGNet with a simple learnable parameter in Net-2 results in inferior performance, highlighting the importance of carefully designing CGNet to enhance the representation ability of the sparse model. In Net-3, where $\bm W$ is fixed and unlearnable, the reconstruction quality suffers due to the limited adaptability of pre-defined sparsifying frames, which fail to capture data-specific structural nuances. In Net-4, introducing tight constraints on $\bm W$ restricts its flexibility, leading to suboptimal performance compared to the unconstrained $\bm W$ in Net-5. By contrast, Net-5 leverages the deep unfolding technique, allowing $\bm W$ to be learned end-to-end in a relaxed and adaptive manner. This enables LipNet to achieve superior results by maximizing the representation ability of sparsifying frames $\bm W$ while avoiding the restrictive nature of tight frame constraints.\par
\begin{figure}[!h]
		\centerline{\includegraphics[width=0.9\linewidth]{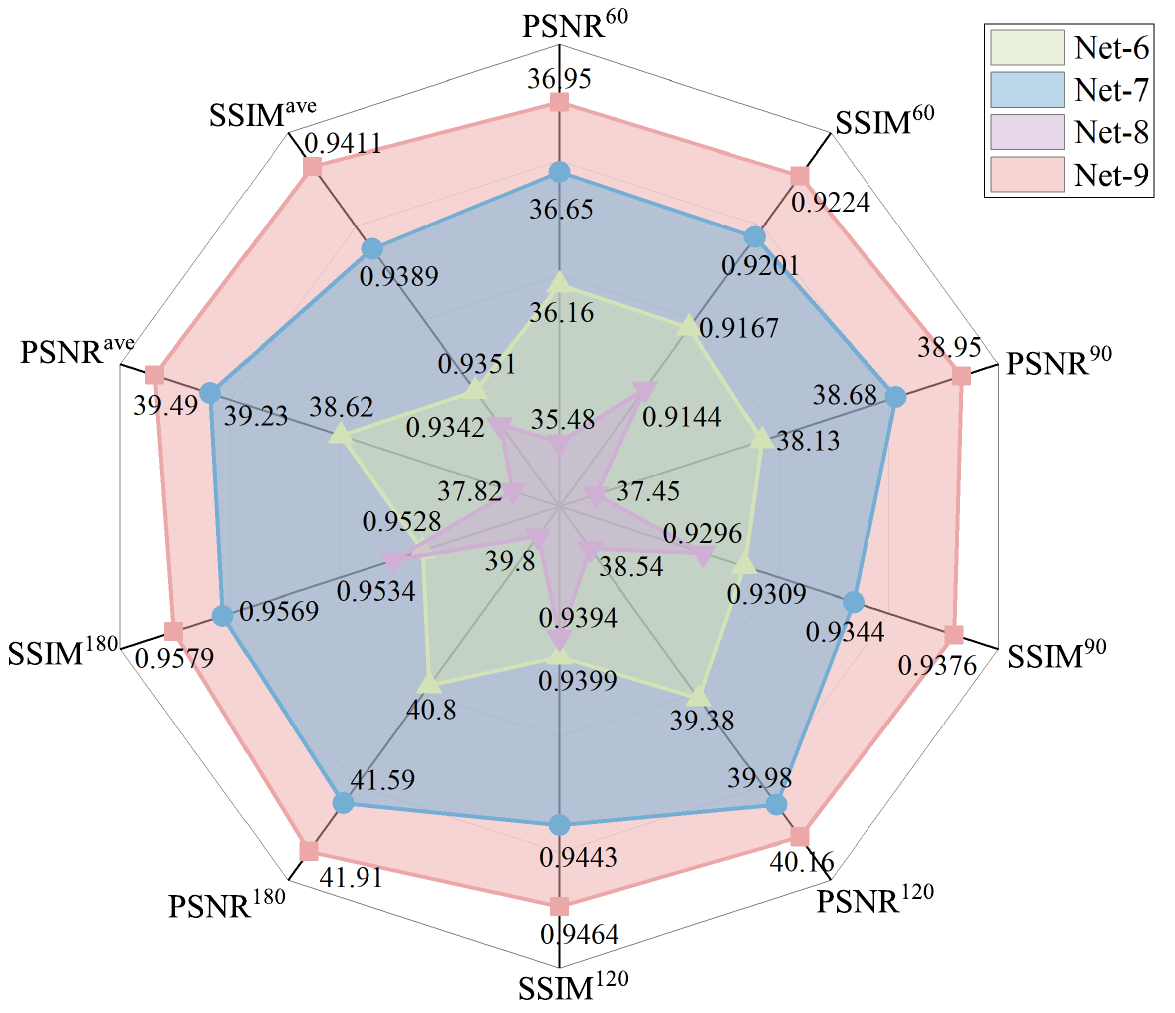}}
		\caption{Comparison of PSNR (dB) and SSIM for different prompt learning on the AAPM dataset under various SVCT scanning settings. 60, 90, 120, and 180 are the sparse view numbers we choose and their average scores are calculated. In radar charts, the farther away from the center, the higher the value, indicating better SVCT reconstruction results.}
		\vspace{-0.4cm}
		\label{fig8}
\end{figure}
\begin{figure}[!h]
	\centerline{\includegraphics[width=0.9\columnwidth]{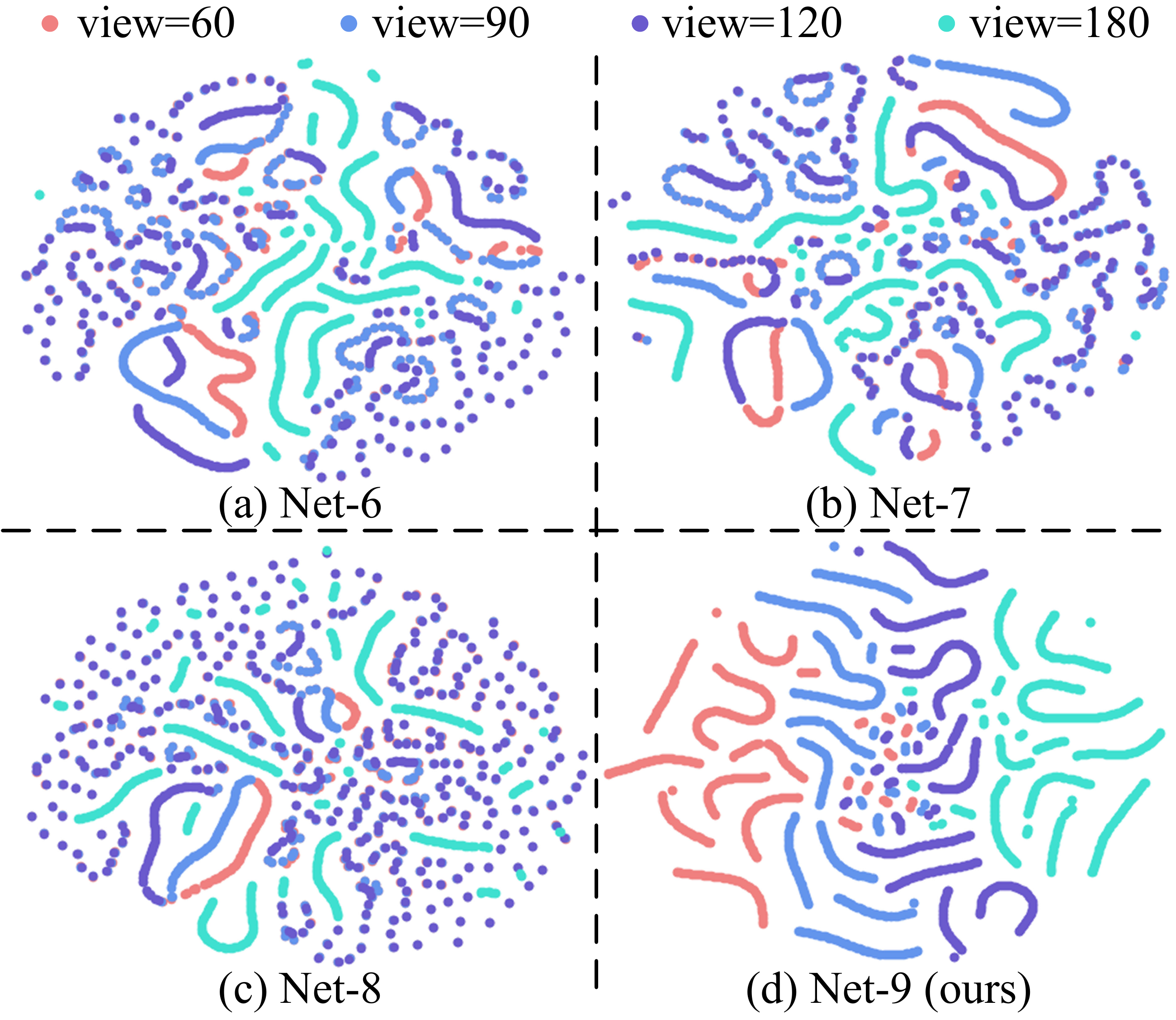}}
	\caption{Visualization of the features of prompt learning using t-SNE. The same object is coded by the same color. This indicates that most of features under explicit prompts are distinguishable, but the features become indistinct under adaptive prompts or feature-learning prompts conditions.}
	\vspace{-0.4cm}
	\label{fig9}
\end{figure}
\subsubsection{The effect of explicit prompts}
\label{subsec:The effect of explicit prompts}
We conduct ablation experiments on explicit view prompts based on the PromptCT architecture. The configurations are as follows: (\textit{\romannumeral1}) Net-6: removing view prompts from the default; (\textit{\romannumeral2}) Net-7: replacing explicit view prompts with adaptive prompts\cite{38}; (\textit{\romannumeral3}) Net-8: replacing explicit view prompts with feature-learning prompts\cite{39}; (\textit{{\romannumeral4}}) Net-9: default version with explicit prompts. The quantitative comparisons of different prompt learning methods are shown in Fig. \ref{fig8}. We observe that both adaptive prompts and feature-learning prompts are capable of perceiving different sparse view settings. However, due to the limited size of the training dataset, these pieces of information are not fully exploited for effectively distinguishing fine-grained settings, resulting in suboptimal performance. In contrast, the proposed explicit view prompts benefit our model by injecting downsampling sinogram masks to distinguish sparse sampling information. As illustrated in Fig. \ref{fig9}, we can see that the explicit view prompts are more capable of distinguishing specific sparse-view information than the adaptive prompts and the feature-learning prompts to encode the key discriminative information of artifact-affected CT images. The explicit prompt module can incorporate view prompts into feature information to distinguish different sparse sampling configurations, thereby providing decoupled properties for the multiple-in-one model and simultaneously enhancing reconstruction performance. Due to the inherent variability and adaptability of the learning process, explicit prompt-based learning methods often exhibit subtle feature overlaps, especially when sparse view angles are numerically close. It is important to note that in prompt learning frameworks, minor feature overlaps or mixing are expected and acceptable, as they reflect the inherent flexibility of the model in capturing different patterns.
\subsubsection{The effect of STB/SFB in CGNet}
\label{subsec:The effect of STB/SFB in CGNet}
\begin{table}[!h]
	\centering
	\caption{Ablation study on the effect of STB/SFB in CGNet on AAPM dataset at the sparse view number of 120.}
	\footnotesize{\tabcolsep=10pt}
	\resizebox{0.8\linewidth}{!}{
		\begin{tabular}{cccc}
			\toprule[1.3pt]
			\textbf{Network}   &\textbf{STB} &\textbf{SFB} &\textbf{PSNR $\backslash$ SSIM $\backslash$ RMSE}\\
			\midrule	
			Net-10           &$\times$    &$\times$   &37.53 $\backslash$ 0.9298 $\backslash$ 0.0134        \\
			Net-11            &\checkmark    &$\times$    &39.34 $\backslash$ 0.9409 $\backslash$ 0.0109      \\
			Net-12            &$\times$     &\checkmark    &39.77 $\backslash$ 0.9441 $\backslash$ 0.0104      \\
			Net-13            &\checkmark     &\checkmark    &40.10 $\backslash$ 0.9466 $\backslash$ 0.0100       \\
			\bottomrule[1.3pt]
	\end{tabular}}
%	\vspace{-0.3cm}
	\label{tab3}
\end{table}
To evaluate the individual contributions of STB and SFB within CGNet, we conduct ablation experiments, as summarized in Tab. \ref{tab3}. The results indicate that removing either STB or SFB significantly reduces the reconstruction quality, as these components play critical roles in the network. STB enhances global contextual modeling by capturing long-range dependencies, while SFB focuses on extracting fine-grained spatial frequency features essential for preserving structural details. Without these modules, the network struggles to suppress artifacts and maintain structural integrity, leading to poorer reconstruction performance. Combining both modules achieves the best results, showcasing their complementary strengths in addressing SVCT challenges.\par
\subsection{Convergence and stability}
\label{subsec:Convergence validation}
\begin{figure}[!h]
	\centering
	\includegraphics[width=1\linewidth]{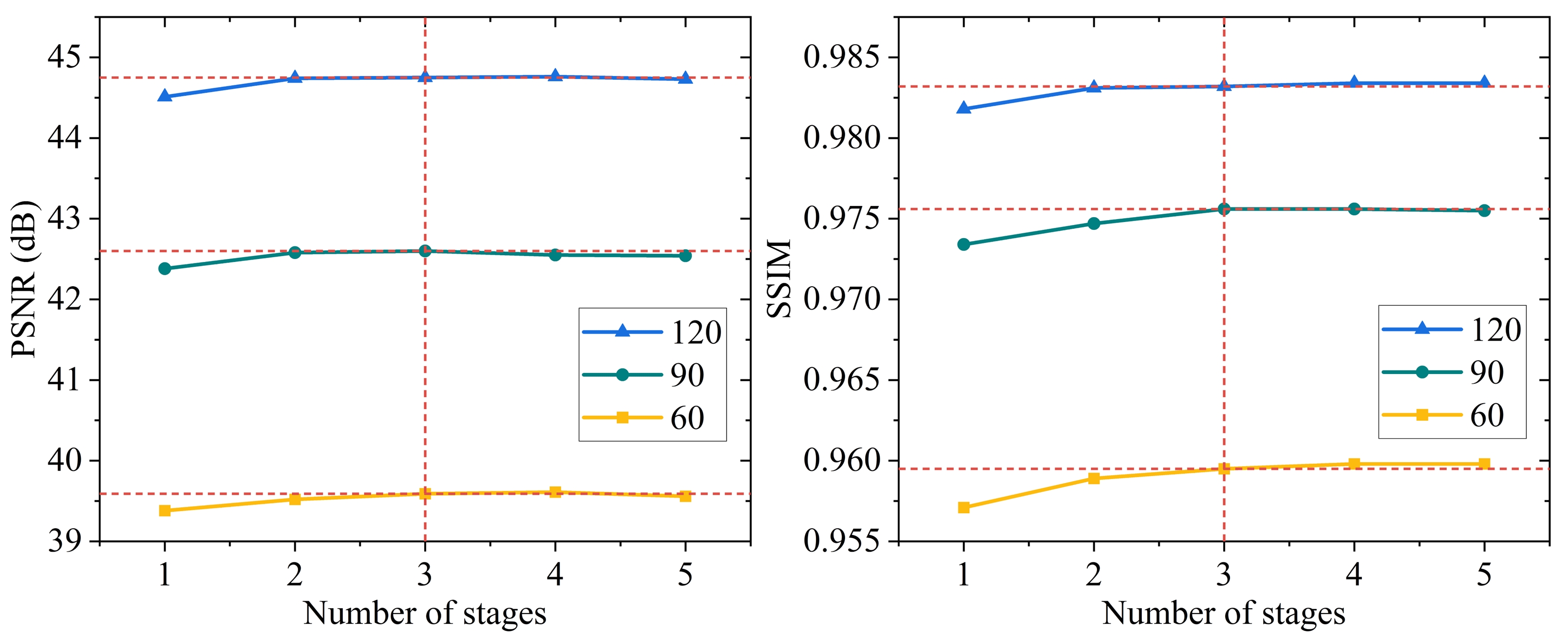}
	\caption{Average PSNR (dB)$\color{red}{\uparrow}$/ SSIM$\color{red}{\uparrow}$ values for increasing network stages $K=1,2,3,4,5$ in the inference process on the DeepLesion dataset under different sparse-view settings.}
	\vspace{-0.4cm}
	\label{fig12}
\end{figure}
\begin{figure}[!h]
	\centering
	\includegraphics[width=1\linewidth]{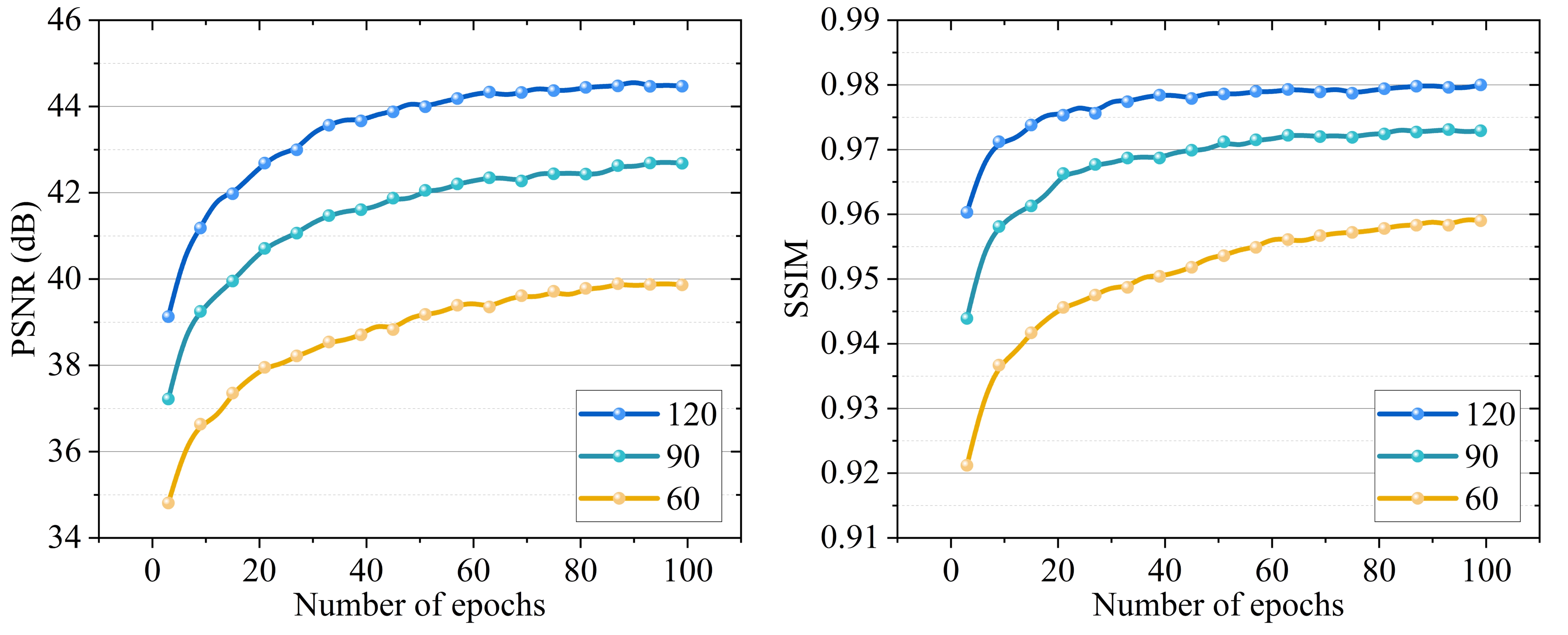}
	\caption{Average PSNR (dB)$\color{red}{\uparrow}$/ SSIM$\color{red}{\uparrow}$ values for increasing epochs in the training process on the DeepLesion dataset under different sparse-view settings.}
	\vspace{-0.2cm}
	\label{fig16}
\end{figure}
To experimentally validate the convergence property of our proposed iterative algorithm, we plot the reconstruction quality metrics (i.e., PSNR and SSIM) of reconstructed CT images at $N_{p} = 60, 90, 120$ for various numbers of stages (i.e., $K = 1, 2, 3, 4, 5$). As shown in Fig. \ref{fig12}, the proposed algorithm demonstrates stable and rapid convergence, with reconstruction accuracy steadily improving as the number of stages increases. Specifically, the average PSNR, SSIM, and RMSE values converge around $K=3$, and the improvement in reconstruction performance gradually levels off as the number of stages increases. Theoretically, adding more stages enhances reconstruction quality by providing deeper iterative refinements. However, this improvement comes at the cost of increased computational time and complexity, resulting in lower computational efficiency. Considering the trade-offs between reconstruction quality and computational cost, we ultimately determined $K=3$ to be the optimal number of stages for training. Furthermore, we provide a practical visualization of the training process. Figure \ref{fig16} illustrates the reconstruction performance in terms of PSNR and SSIM as functions of the number of epochs under three different sparse-view conditions (i.e., $N_{p} = 60, 90, 120$). The results indicate that higher sparse sampling rates achieve better reconstruction quality, with PSNR and SSIM values steadily increasing and stabilizing as the number of epochs progresses, demonstrating the convergence of our method during the training phase.\par
\subsection{Feasibility of 3D reconstruction}
\label{subsubsec:3D image volume reconstruction}
\begin{figure}[!h]
	\centering
	\includegraphics[width=0.8\linewidth]{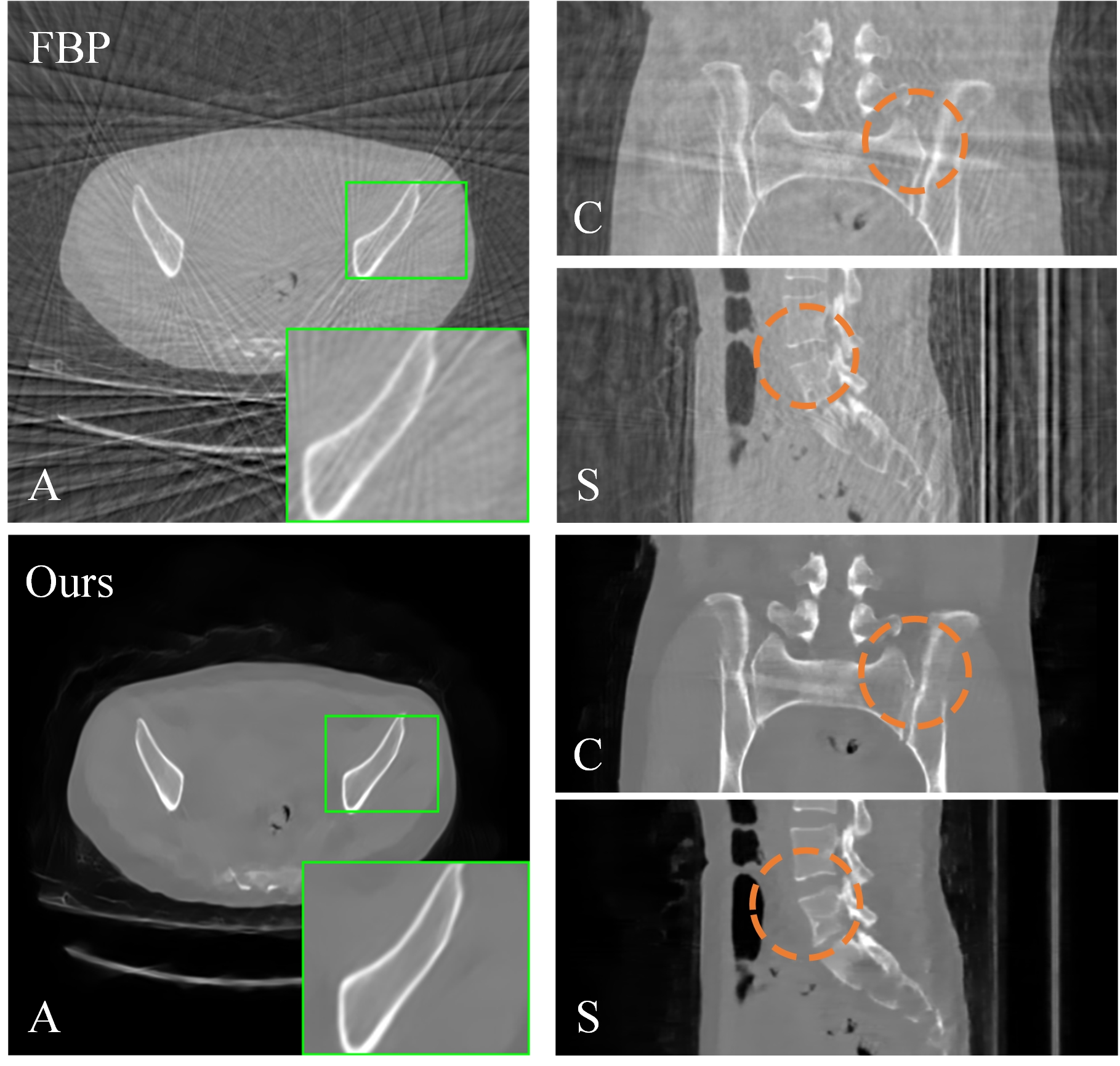}
	\caption{The reconstruction results on the SpineWeb dataset at the sparse view number of 60. The 3D image volume is obtained by stacking all the axial slices together. A: Axial, C: Coronal, S: Sagittal. The green rectangles highlight regions of interest, which are zoomed in for better viewing. Orange circles indicate the hip and lumbar spine in the CT images.}
	\vspace{-0.2cm}
	\label{fig14}
\end{figure}
To validate the extension of our method to the reconstruction of 3D image volumes, we utilize CT slices from a patient in the SpineWeb dataset at the sparse view number of 60, specifically focusing on vertebrae localization and identification. As shown in Fig. \ref{fig14}, we generate axial, coronal, and sagittal CT slices from the SpineWeb dataset and apply our proposed method to reconstruct clear and detailed images across these orientations. The 3D image volume is constructed by stacking all the axial slices together. The analytical reconstruction algorithm, FBP, struggles with incomplete projection data, resulting in severe streak artifacts in the reconstructed slices. In contrast, our method demonstrates significant improvements, effectively removing artifacts and restoring subtle structures. The results indicate that our approach achieves stable reconstruction accuracy across all orientations, preserving clear details of the vertebrae and surrounding bones. This demonstrates the robustness of our method in producing high-quality reconstructions in multiple dimensions.\par
\section{Conclusion and Future work}
\label{sec:Conclusion}
In this work, we proposed a storage-saving deep unfolding framework for SVCT reconstruction, called PromptCT. Compared to existing models, we employed a single model to address SVCT reconstruction setups across various sparse sampling view numbers, thus significantly alleviating the storage cost of relevant medical institutions. On the theoretical side, we explicitly dropped the requirement for the closed-form solution and proposed a prompt-based sparse representation model-driven network satisfying Lipschitz constraint. On the experimental side, PromptCT achieved competitive or even better reconstruction performance at lower storage costs in multiple-in-one SVCT reconstruction compared with the benchmark methods, demonstrating the effectiveness and applicability of our proposed model.\par
In future work, we aim to address the challenges of GPU memory consumption and extend the proposed network to 3D reconstruction tasks by optimizing its computational efficiency and exploring lightweight architectures. Additionally, we will incorporate non-linear measurement models and validate the network on diverse real-world datasets, further bridging the gap between simulation and practical CT reconstruction applications. These efforts will enhance the scalability and adaptability of our network, making it suitable for broader clinical and industrial scenarios. Furthermore, we will use larger datasets and incorporate more clinically relevant evaluation metrics. These improvements will allow for a more robust and clinically applicable method for CT reconstruction.\par

\bibliography{ref.bib}
\end{document}